\documentclass{article}
\usepackage{iclr2027_conference,times}
\usepackage[T1]{fontenc}
\usepackage{amsmath,amssymb,amsthm}
\usepackage{siunitx}
\usepackage{graphicx,adjustbox}
\usepackage{array,tabularx,booktabs,makecell,multirow}
\usepackage{xcolor}                 
\usepackage{enumitem,capt-of,titlesec}
\usepackage[most]{tcolorbox}
\usepackage{placeins}
\usepackage{booktabs}

\usepackage{amsmath,amsfonts,bm}

\def\eqref#1{equation~\ref{#1}}

\def\1{\bm{1}}

\DeclareMathAlphabet{\mathsfit}{\encodingdefault}{\sfdefault}{m}{sl}
\SetMathAlphabet{\mathsfit}{bold}{\encodingdefault}{\sfdefault}{bx}{n}

\renewcommand{\eqref}[1]{\textup{(\ref{#1})}}

\colorlet{gainpos}{green!45!black}
\colorlet{gainneg}{red!70!black}
\colorlet{gainzero}{black!40}
\newlength{\gainwd}
\newcommand{\gainsub}[2]{\textsubscript{\makebox[\gainwd][l]{\textcolor{#1}{#2}}}}
\newcommand{\gup}[1]{\gainsub{gainpos}{+#1}}
\newcommand{\gdn}[1]{\gainsub{gainneg}{$-$#1}}
\newcommand{\gz}{\gainsub{gainzero}{\,0.0}}

\theoremstyle{definition}
\newtheorem{theorem}{Theorem}[section]
\newtheorem{proposition}[theorem]{Proposition}

\newtheoremstyle{appendixplain}
  {5pt}
  {2pt}
  {\itshape}
  {0pt}
  {\bfseries}
  {.}
  {0.5em}
  {\thmname{#1}\thmnumber{ #2}\thmnote{ (#3)}}
\theoremstyle{appendixplain}

\newtheorem{applemma}[theorem]{Lemma}

\newtheorem{appcorollary}[theorem]{Corollary}
\newcommand{\restatedname}{}
\newtheorem*{restatedthm}{\restatedname}
\newenvironment{restated}[3]
  {\renewcommand{\restatedname}{#1~\ref{#2}}\begin{restatedthm}[#3]}
  {\end{restatedthm}}
\newtcolorbox{matchedexample}[1]{
  colback=black!2,
  colframe=black!35,
  colbacktitle=black!7,
  coltitle=black,
  title=\textbf{#1},
  fonttitle=\small\bfseries,
  fontupper=\small,
  boxrule=0.4pt,
  arc=0pt,
  left=5pt,
  right=5pt,
  top=4pt,
  bottom=4pt,
  before skip=4pt,
  after skip=4pt
}

\usepackage{hyperref}
\hypersetup{pdfauthor={}}
\usepackage{url}

\title{Rethinking Fixed-Horizon Latent Reasoning}
\author{Tianxiang Chen, Jingyuan Zhou, Weiming Yang, Kaidi Yang\thanks{Corresponding author.} \\
Department of Civil and Environmental Engineering \\
National University of Singapore \\
\texttt{\{e1373675,jingyuanzhou\}@u.nus.edu} \\
\texttt{\{weiming\_yang,ykaidi\}@nus.edu.sg}}

\iclrfinalcopy
\begin{document}
\maketitle
\begin{abstract}
Existing latent reasoners typically run for a preset latent horizon during inference. Intuitively, a longer horizon should provide more opportunity for reasoning. However, in this work, we find that extending latent Chain-of-Thought (CoT) to the full preset horizon may in fact degrade reasoning performance. We observe that latent CoT evolves through distinct functional stages, with earlier computation forming frontier reasoning states, while later computation mainly reorganizes existing candidates and can weaken gold candidate support. These findings motivate a rethink of fixed-horizon latent inference. To address this, we characterize the reasoning benefits of latent CoT by realized and remaining constructive progress. Empirically, we find that higher realized constructive progress corresponds to higher current reasoning accuracy, while more remaining progress corresponds to larger future gains from continued latent CoT. To this end, we introduce a lightweight method that estimates the remaining reasoning progress and adaptively adjusts the latent horizon during inference. Across representative latent CoT reasoners and multiple arithmetic benchmarks, our method improves reasoning accuracy by up to $7.2$ percentage points while reducing latent computation by up to $29.0\%$, without retraining the LLM.
\end{abstract}

\section{Introduction}
Recent advances in latent reasoning~\citep{deng2023icot,hao2025coconut} have established latent Chain-of-Thought (CoT) as an alternative to explicit CoT reasoning~\citep{wei2022cot}. Central to the appeal of latent CoT is its ability to carry intermediate reasoning directly in continuous representations, without realizing each step as discrete language tokens~\citep{xu2025softcot,zhang2025softthinking}. Prior work has demonstrated encouraging capabilities of this paradigm, including the maintenance of multiple reasoning alternatives~\citep{zhu2025superposition} and the compression of explicit reasoning into continuous states~\citep{cheng2024ccot,shen2025codi}. Existing latent reasoners typically perform a fixed number of latent steps before generating an explicit answer~\citep{wei2026simcot}. Intuitively, a longer latent horizon should provide more opportunity for latent CoT to advance its reasoning~\citep{goyal2024pause}. Yet, while prior work has established what latent CoT can achieve~\citep{tan2025colar}, how its reasoning benefits evolve across latent steps remains poorly understood.

\begin{figure}[htbp]
    \centering
    \includegraphics[width=1\linewidth]{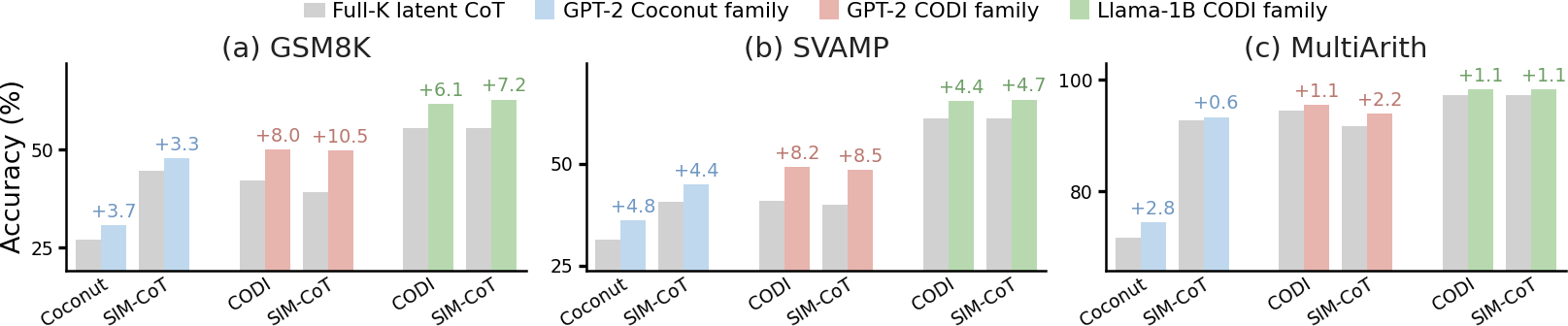}
    \caption{\textbf{The full latent horizon can bury better reasoning outcomes.}
    We evaluate six trained latent CoT reasoners across GPT-2~\citep{radford2019gpt2} and Llama-1B~\citep{grattafiori2024llama3} backbones, covering Coconut, CODI, and SIM-CoT variants. For each problem, we retrospectively stop latent reasoning at the first depth that produces a correct answer, if such a depth exists. Gray bars show accuracy at the preset full horizon, while colored bars show accuracy under this retrospective stopping point.}
  \label{fig:1}
\end{figure}

In this work, we find that completing the full latent horizon may in fact degrade reasoning performance. To assess this, we compare the outcome at the preset full horizon with the best outcome reached earlier along the same latent trajectory. As shown in Figure~\ref{fig:1}, better reasoning outcomes can emerge before the preset horizon is exhausted and fail to be preserved at the full horizon. This suggests that reasoning benefits do not simply accumulate as latent CoT becomes deeper.

This key observation is closely related to how latent CoT evolves along the reasoning trajectory. We find that the latent trajectory can exhibit three distinct stages: frontier advancement, candidate organization, and answer revision. At the frontier stage, earlier evolution continues to form new task-relevant reasoning states, while later organization increasingly reorganizes candidates that have already emerged. As answer candidates become established, later latent steps mainly revise existing candidates and yield diminishing reasoning gains as latent evolution continues. Thus, the reasoning value of latent CoT is non-uniform along the reasoning trajectory.

The above observations motivate a rethink of fixed-horizon latent inference.  Fixed-horizon inference executes the same preset depth regardless of how reasoning has progressed along the latent trajectory. Ideally, latent CoT should continue while further computation still benefits the reasoning process. Once later latent steps mainly revise gold candidates with diminishing additional gains, computation should be curtailed~\citep{chen2024overthinking}. However, this leaves a key challenge in deciding when to curtail latent CoT, before the gains from future latent steps become observable.

To this end, we introduce a lightweight method that estimates the remaining reasoning progress and adaptively adjusts the latent horizon during inference. Specifically, we estimate how much reasoning progress remains from the observed latent trajectory and use this estimate to determine the realized latent horizon. This allows the executed latent horizon to vary with the reasoning progress of the current trajectory. We evaluate trained latent CoT reasoners spanning CODI~\citep{shen2025codi}, Coconut~\citep{hao2025coconut}, and SIM-CoT~\citep{wei2026simcot} variants on multiple arithmetic benchmarks. Empirically, our method improves final reasoning accuracy across all settings, with gains of up to $7.2$ percentage points while reducing latent computation by up to $29.0\%$. Interested readers can refer to Appendix~\ref{app:related-work} for a thorough discussion of related work.

\textbf{Statement of Contribution}. Our contributions are threefold. First, we show that completing the full latent horizon can degrade reasoning performance and reveal three distinct stages of latent CoT. Second, we formulate latent continuation in terms of remaining reasoning progress and introduce a lightweight method that can adapt the latent horizon during inference. Third, we validate adaptive latent inference on multiple arithmetic benchmarks across CODI, Coconut, and SIM-CoT latent CoT reasoners, consistently improving reasoning accuracy while reducing latent computation.

\section{Problem Formulation}
\textbf{Latent CoT.} Latent CoT carries the intermediate computations as a sequence of latent reasoning states. We index latent computation by $t\in\{1,\ldots,K\}$, where $K$ denotes the preset latent horizon, and let $z_t$ denote the reasoning state after the $t$-th latent step. Let $x$ denote an input problem with gold answer $y^\star$, and let $y=(y_1,\ldots,y_L)\in\mathcal{Y}$ denote a generated answer. Given $x$ and the preceding latent states $z_{<t}$, the model induces $z_t \sim \mathcal{T}_\theta(\cdot\mid x,z_{<t})$. After $t$ latent steps, the model induces an answer distribution $p_\theta(y\mid x,z_{1:t})$. We use a task reward $r(x,y)$ to score answers generated at each depth. In our reasoning tasks, the reward is instantiated as answer correctness with respect to $y^\star$; implementation details are provided in Appendix~\ref{app:Experimental Details-1}. 

\textbf{Adaptive Latent Horizon.} Standard latent CoT inference executes all $K$ preset latent steps before answer generation. Reasoning outcomes can vary across latent depths. In particular, an intermediate latent depth can yield a better reasoning outcome than the full horizon. This observation motivates adjusting the latent CoT budget during inference. Let $\tau\in\{1,\ldots,K\}$ denote the realized latent horizon, with the resulting answer generated as $y\sim p_\theta(\cdot\mid x,z_{1:\tau})$. At step $t$, continuation can depend only on the input $x$ and the prefix latent states $z_{1:t}$. We study how to determine $\tau$ from the latent computation available during inference.

\section{Functional Evolution in Latent Reasoning}
\label{sec:3}

Explicit CoT externalizes intermediate reasoning as a visible sequence of tokens~\citep{nye2021scratchpad}. By contrast, continuous latent reasoning carries intermediate reasoning without explicitly decoding each step, allowing multiple reasoning alternatives to remain unresolved~\citep{gozeten2026cot2}. \textit{However, how these alternatives evolve in the latent space remains unclear}. To examine this, we use reasoning candidates as a common reference and compare their support at successive steps.

\subsection{Structured Progression along Latent Reasoning}

To ground our investigation, we use ProsQA~\citep{hao2025coconut}, where each problem is paired with an explicit reasoning graph. This structure allows us to associate reasoning candidates with different depths. We mainly analyze the symbolic Coconut model~\citep{zhu2025superposition}. Since intermediate reasoning remains latent, directly characterizing its evolution is difficult. Inspired by \citet{hao2025coconut}, we decode reasoning candidates at each latent step and track how their probabilities change. For clarity, we present a selected example here and defer additional results to Appendix~\ref{app:Experimental Details-3}.

\begin{figure}[htbp]
  \centering
  \includegraphics[
    width=1\columnwidth,
    keepaspectratio]{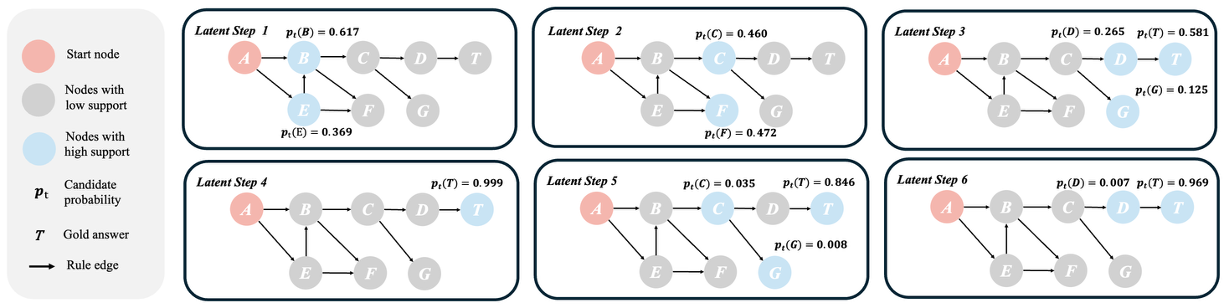}
  \caption{\textbf{An example of latent reasoning progress.} We use the symbolic Coconut model from \citet{zhu2025superposition} and roll it out for six latent steps to trace the reasoning progress. We report the top-6 candidates at each step, with highlighted nodes indicating high support.}
  \label{fig:3.1}
\end{figure}

\textbf{Frontier advancement.} Figure~\ref{fig:3.1} (Steps 1--4) shows the support assigned to reasoning candidates at different graph depths over successive latent steps. Candidates at smaller graph depths receive support earlier, followed by candidates at larger depths, while answer candidates emerge later. Earlier intermediate candidates tend to receive support only briefly, whereas later candidates remain supported across more latent steps. Together, these patterns show an advancing reasoning frontier, with support moving through intermediate reasoning before answer candidates become established.

\textbf{Candidate organization.} As latent reasoning proceeds, as shown in Figure~\ref{fig:3.1} (Steps 3--5), new intermediate reasoning candidates enter the high support region less frequently. In contrast, candidates that have already emerged continue receiving support, while answer candidates increasingly dominate the support. This suggests that latent reasoning is no longer continually forming new candidates with high support, but organizing candidates that have already emerged. We term this stage candidate organization.

\textbf{Answer revision.} Further, as shown in Figure~\ref{fig:3.1} (Steps 4--6), once the gold answer candidate becomes the most supported, later latent steps still shift support among other candidates. In our tests, this shift can weaken the gold candidate's support and even cause it to lose the lead. Meanwhile, few new candidates enter, with the remaining changes concentrated among already formed answer candidates. Thus, we argue that the later latent reasoning enters an answer revision stage: the model mainly reorganizes the support among existing answer candidates, and this revision can overturn an already established gold answer.

\subsection{Reasoning Gains Along the Progression}

Building upon the progress identified above, we empirically test whether latent computation consistently yields further gains. Intuitively, while the solution remains unresolved, latent computation should continue to advance the reasoning frontier, repeatedly updating the top-$k$ candidates. Once the gold answer begins to form, further computation should increasingly consolidate support around a stable solution. To make this intuition concrete, we consider two empirical implications: (1) While the reasoning frontier is still advancing, prematurely realizing the current latent state should reduce subsequent progress toward the gold answer. (2) Once the gold answer has formed, later latent computation should consolidate the gold result if it yields further gains.

\begin{figure}[htbp]
  \centering
  \includegraphics[
    width=1\columnwidth,
    keepaspectratio]{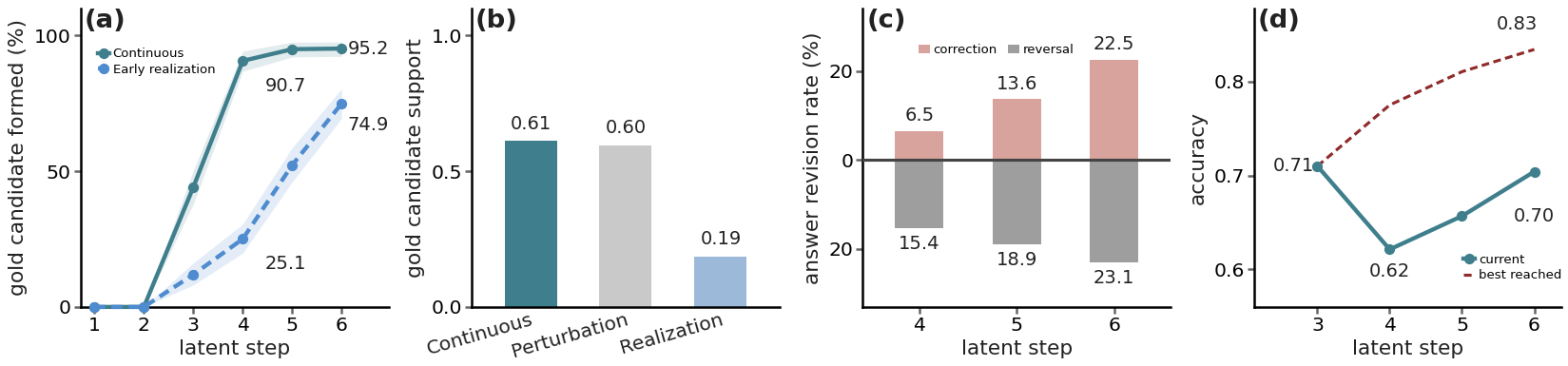}
  \caption{\textbf{Latent progression and revision.}
    \textbf{(a)} Gold candidate formation; \textbf{(b)} Gold candidate support.
    For trajectories whose answer is formed at the third latent step, \textbf{(c)} Correction and reversal rates, and \textbf{(d)} Accuracy over revision vs. the best accuracy reached along the trajectory.}
  \label{fig:3.2}
\end{figure}

To test the first empirical implication, we interrupt the ongoing latent CoT at the frontier stage by inserting a single explicit realization. Specifically, we decode the current state to its top token, add this token to the causal context, and continue latent computation. As shown in Figure~\ref{fig:3.2}(a), realization markedly delays subsequent progression toward the gold answer. To more rigorously assess the gain during the frontier stage, we introduce a matched control that perturbs the continuous state by a comparable magnitude. As shown in Figure~\ref{fig:3.2}(b), with perturbations, subsequent candidate progression remains largely preserved. Inspecting the top-$k$ candidates further, we find that the gold answer already appears while the frontier is still advancing, suggesting that frontier stage gains can continue after the answer first emerges. Taken together, these results suggest that the function of the frontier stage is to sustain candidate progression toward the gold answer, thereby yielding further reasoning gains.

To examine the second empirical implication, we assess whether later latent computation continues to consolidate the gold result after it has emerged. We quantify this by tracking later corrections toward the gold answer and reversals of an already established gold answer. As shown in Figure~\ref{fig:3.2}(c), later computation recovers the gold answer in some cases, while also overturning gold results that have already been established, leaving little net gain. This suggests that later revisions yield little stable additional reasoning gain. More importantly, as illustrated in Figure~\ref{fig:3.2}(d), some later revisions overturn an already established gold result, allowing gains reached earlier in the trajectory to be lost during continued computation. Consequently, continued latent evolution can weaken reasoning gains that have already emerged.

Together, these results show that the effect of latent CoT is non-uniform, giving rise to three functional stages. Specifically, frontier advancement continues to yield reasoning gains, candidate organization marks the transition, and later revision mainly reorganizes the formed candidates. In other words, once constructive progress has largely saturated, additional latent computation offers diminishing returns and even becomes detrimental.

\section{Remaining Reasoning Progress in Latent Computation}

So far, we have identified distinct reasoning stages in latent CoT, revealing non-uniform reasoning benefits across latent steps. In this section, we first examine the relation between latent CoT evolution and remaining progress (Section~\ref{sec:4.1}). Building on this, we propose a lightweight method that can estimate the remaining reasoning progress and adaptively adjust the latent horizon during inference to improve latent thinking (Section~\ref{sec:4.2}). Finally, we provide a formal characterization of progress estimation and allocation (Section~\ref{sec:4.3}).

\subsection{Reasoning Gains Beyond the Current Frontier}
\label{sec:4.1}
The stage structure suggests a simple strategy: continue latent CoT during the frontier stage and stop once reasoning shifts to revision. This strategy, however, introduces two difficulties.

\textbf{Ambiguity in latent gains.} Intuitively, the extent of latent evolution may appear to provide a natural signal of reasoning progression~\citep{geiping2026scaling}. Yet, we find that substantial latent evolution can persist during revision, making state evolution insufficient for distinguishing different stages. Moreover, the same ambiguity persists in realized reasoning gains, since a substantial local gain does not necessarily establish a new frontier. For instance, later revision can induce large fluctuations that recover a previously attained value without advancing the frontier.

\textbf{Dependence on future gains.} Section~\ref{sec:3} shows that latent evolution can persist even as genuinely new reasoning gains become sparse. Adapting the budget of latent CoT therefore requires judging whether further computation can still produce additional gain, which is realized in the remaining latent suffix $z_{t+1:K}$, while inference only has access to the realized computation $z_{1:t}$. The appropriate endpoint can therefore only be identified retrospectively from future gains, whereas continuation must be decided online from the current state.

\subsection{From Remaining Progress to an Adaptive Latent Horizon}
\label{sec:4.2}
To resolve the ambiguity, we evaluate each latent depth against the best reasoning outcome reached so far, which defines the current reasoning frontier. Then, a latent step can establish constructive progress only when it advances this frontier. For online continuation, the core question is whether latent CoT is worth continuing. We address this in two steps: estimating how much reasoning progress remains and adapting the latent horizon accordingly.

\textbf{Formulation.} Consider an input $x$ and a $K$-step latent CoT $z_1,\ldots,z_K$. After $t$ latent steps, we define the current reasoning value as 
\begin{equation}
 G_t = \mathbb{E}_{y\sim p_\theta(\cdot\mid x,z_{1:t})} \bigl[r(x,y)\bigr]. 
 \label{eq:reasoning_value}
\end{equation}
To quantify the remaining reasoning progress, we compare the reasoning frontier reached up to step \(t\) with the best frontier attainable within the full latent horizon, written as:
\begin{equation}
\Omega_t
=
\max_{1\le j\le K} G_j
-
\max_{1\le j\le t} G_j .
\label{eq:remaining_progress}
\end{equation}

By this definition, we now assess whether latent reasoning is worth continuing by estimating how much further frontier advancement remains. At inference, $\Omega_t$ is not directly available because it depends on future reasoning gains. To estimate it from the current latent prefix, we represent the observed evolution of latent CoT with a causal trajectory representation $X_t$, which captures short-range changes in its predictive statistics and observed hidden state. We defer its detailed construction to Appendix~\ref{app:Experimental Details-2}.

\textbf{Estimation.}
The above formulation allows us to train a lightweight estimator $f_\phi$ to predict the remaining progress from $X_t$. To construct the training targets, we run the fixed latent reasoner over the full $K$ step trajectory and roll out from each checkpoint to estimate its reasoning value $\hat{G}_t$, yielding the remaining progress target $\hat{\Omega}_t$ according to Eq.~\eqref{eq:remaining_progress}. We optimize $f_\phi$ by minimizing the squared error between prediction and the remaining progress target:

\begin{equation}
\mathcal{L}_{\mathrm{prog}}(\phi)
=
\mathbb{E}_{(X_t,\widehat{\Omega}_t)}
\left[
\left(
f_\phi(X_t)-\widehat{\Omega}_t
\right)^2
\right].
\label{eq:progress_estimator}
\end{equation}

\textbf{Horizon adaptation.} With this estimate, identifying an exact stopping point would still require predicting when future progress will occur. We aim for a common continuation mechanism across latent reasoners, even when progress is represented differently across backbones. Accordingly, we express each estimated remaining progress by its relative position in a frozen reference distribution. For each latent reasoner, let $F_{\mathrm{ref}}$ denote the cumulative distribution of its reference progress estimates. We define
\begin{equation}
u_t = F_{\mathrm{ref}}\!\left[f_\phi(X_t)\right].
\label{eq:relative_progress}
\end{equation}
Thus, $u_t$ represents the relative continuation demand associated with the estimated remaining progress. We use $u_t$ to control how much additional latent computation is allocated over the remaining horizon. Let $a_t$ denote the additional latent computation allocated after step $t$. We write
\begin{equation}
a_t = (K-t)\,g(u_t),
\label{eq:latent_allocation}
\end{equation}
where $g$ is a bounded, non-decreasing continuation rule, with its concrete form deferred to Appendix~\ref{app:Experimental Details-2}. After discretization, the allocation yields the realized horizon $\tau$. In this way, we replace the preset horizon $K$ with an adaptive horizon determined solely from the current state.

\subsection{Latent Trajectories and Returns from Additional Computation}
\label{sec:4.3}
One might wonder whether the latent trajectory provides sufficient information to adapt the remaining latent budget. To understand this, we examine how the latent CoT supports remaining progress estimation and how remaining progress affects the additional latent gain.

\textbf{Estimation.} Although $\Omega_t$ depends on later reasoning gains, online continuation only has access to $X_t$. We therefore characterize the optimal prediction of $\Omega_t$ from $X_t$ under population squared risk.

\begin{proposition}[Causal Estimation of Remaining Progress]
\label{prop:causal-estimation}
\emph{Under the population squared risk
$\mathcal{R}(f)=\mathbb{E}[(f(X_t)-\Omega_t)^2]$,
the optimal estimator of remaining progress from the latent trajectory satisfies}
\begin{equation}
f^*(X_t)=\mathbb{E}\!\left[\Omega_t\mid X_t\right].
\label{eq:optimal_estimator}
\end{equation}
\end{proposition}
The proof is provided in Appendix~\ref{app:causal_estimation}. Thus, remaining progress can be predicted from the latent trajectory through its conditional expectation, which defines the population target learned by \(f_\phi\).

\textbf{Allocation.} For continuation, we consider how the reasoning gain from additional latent computation varies with remaining progress. Let $\omega$ denote the remaining progress and $a$ the additional latent computation. We define $V(a,\omega)$ as the expected change in reasoning value after $a$ additional latent steps when the remaining progress is $\omega$. We consider the condition that a larger $\omega$ yields a larger marginal gain from additional latent computation. At latent depth $t$, let $A=K-t$ denote the maximum additional latent budget.

\begin{theorem}[Monotone Allocation by Remaining Progress]
\label{thm:monotone-allocation}
\emph{Let $V(a,\omega)$ have increasing differences in $(a,\omega)$. For trajectories $i=1,\ldots,n$ with remaining progress $\omega_i$, consider integer allocations $a_i\in\{0,\ldots,A\}$ that maximize $\sum_i V(a_i,\omega_i)$ subject to a shared compute budget $\sum_i a_i \leq B$. Then there exists an optimal allocation $a^*$ such that $\omega_i > \omega_j$ implies $a_i^* \geq a_j^*$.}
\end{theorem}

Under these conditions, trajectories with more remaining progress receive no less additional latent computation. We defer the proof to Appendix~\ref{app:monotone_allocation}. Thus, the value of additional latent computation varies systematically with the remaining progress. Together, these results recast latent continuation in terms of online reasoning progress, providing a basis for estimating the remaining progress along the latent trajectory and adapting the latent horizon accordingly.

\section{Experiments}

In this section, we investigate how reasoning progress relates to both the outcomes of latent CoT and the gains from continued reasoning. We evaluate latent CoT reasoners spanning CODI, Coconut, and SIM-CoT variants on GSM8K~\citep{cobbe2021gsm8k}, SVAMP~\citep{patel2021svamp}, and MultiArith~\citep{roy2015multiarith}, with experimental settings provided in Appendix~\ref{app:setup}. Our experimental design aims to examine three claims: (i) that adapting the latent horizon improves final reasoning outcomes while reducing latent computation, (ii) that higher realized constructive progress corresponds to higher current reasoning accuracy, and (iii) that more remaining progress corresponds to larger future gains from continued latent CoT.

\subsection{Adaptive Latent Inference and Reasoning Outcomes}
We first report the overall results of adaptive latent inference on final reasoning accuracy. As shown in Table~\ref{tab:results_main}, adaptive inference achieves higher final accuracy than the original Full-\(K\) inference across all settings. The largest gain is observed on MultiArith with GPT-2/Coconut, where accuracy improves from \(71.7\%\) to \(78.9\%\). Moreover, the improvement persists even in settings where the original Full-\(K\) accuracy is already high and further improvement is more limited. For instance, on MultiArith with Llama-1B/CODI, accuracy still increases from \(97.2\%\) to \(98.3\%\). Meanwhile, adaptive inference also reduces latent computation, with notable savings on latent steps. Consequently, we show that adaptive latent inference enables the same latent reasoner to achieve higher final reasoning accuracy while using less latent computation.

One might wonder whether, once the full preset horizon is no longer necessary, a shorter fixed-horizon would already suffice. To examine this, we compare fixed depth and adaptive inference under the same latent step budgets. As shown in Table~\ref{tab:frontier}, the advantage of adaptive inference is already pronounced even under a limited latent step budget. For example, on GSM8K, the accuracy of GPT-2/CODI increases from \(36.6\%\) to \(41.5\%\) at four latent steps. The advantage persists as the latent CoT depth increases, remaining evident at both five and six steps. Delving into the fixed depth performance, we find that increasing the latent step budget can improve, leave unchanged, or even reduce reasoning accuracy. This pattern is consistent with our observation in Section~\ref{sec:3} that additional latent computation yields distinct gains across three reasoning stages. Thus, these results support adapting the CoT depth according to the reasoning progress throughout inference.

Taken together, these results suggest that continued latent computation can alter whether reasoning gains established earlier are retained in the final prediction. Accordingly, with fixed-horizon inference, latent CoT proceeds to the preset depth even after useful reasoning gains have formed, and later revision can alter the resulting reasoning outcome. By contrast, adaptive latent inference adjusts the CoT depth based on reasoning progress, curtailing computation as further gains diminish. In this way, we can preserve constructive reasoning progress with less latent computation, resulting in better reasoning performance.

\begin{table}[htbp]
    \centering
    \setlength{\abovecaptionskip}{2pt}
    \setlength{\belowcaptionskip}{3pt}
    \setlength{\aboverulesep}{0.3ex}
    \setlength{\belowrulesep}{0.3ex}
    \caption{\textbf{Adaptive latent CoT inference.} Full-\(K\) denotes the preset latent horizon while ours uses adaptive latent inference. Latent step saving (\%) reports the average step reduction.}
    \label{tab:results_main}
    \small
        \begin{tabular}{@{}l@{\hspace{10pt}}c@{\hspace{7pt}}c@{\hspace{7pt}}c@{\hspace{14pt}}c@{\hspace{7pt}}c@{\hspace{7pt}}c@{\hspace{14pt}}c@{\hspace{7pt}}c@{\hspace{7pt}}c@{\hspace{8pt}}c@{}}
        \toprule
        & \multicolumn{3}{c}{GSM8K} & \multicolumn{3}{c}{SVAMP} & \multicolumn{3}{c}{MultiArith} & \multicolumn{1}{c}{Latent step} \\
        \cmidrule(lr){2-4}\cmidrule(lr){5-7}\cmidrule(l){8-10}
        Method & Full-$K$ & Ours & Gain & Full-$K$ & Ours & Gain & Full-$K$ & Ours & Gain & saving (\%) \\
        \midrule
        \multicolumn{11}{@{}l}{\emph{GPT-2}} \\
        \hspace{0.8em}CODI & 42.1 & \textbf{43.4} & +1.3 & 41.0 & \textbf{43.1} & +2.1 & 94.4 & \textbf{95.0} & +0.6 & 20.6 \\
        \hspace{0.8em}SIM-CoT & 39.2 & \textbf{41.5} & +2.3 & 40.0 & \textbf{43.5} & +3.5 & 91.7 & \textbf{92.2} & +0.5 & 29.0 \\
        \addlinespace[2pt]
        \hspace{0.8em}Coconut & 27.1 & \textbf{27.7} & +0.6 & 31.4 & \textbf{33.7} & +2.3 & 71.7 & \textbf{78.9} & \textbf{+7.2} & 11.0 \\
        \hspace{0.8em}SIM-CoT Coconut & 44.7 & 44.7 & +0.0 & 40.6 & \textbf{42.2} & +1.6 & 92.8 & \textbf{93.3} & +0.5 & 25.1 \\
        \cmidrule(r){1-4}\cmidrule(lr){5-7}\cmidrule(l){8-10}\cmidrule(l){11-11}
        \multicolumn{11}{@{}l}{\emph{Llama-1B}} \\
        \hspace{0.8em}CODI & 55.5 & \textbf{55.7} & +0.2 & 61.1 & \textbf{62.3} & +1.2 & 97.2 & \textbf{98.3} & +1.1 & 26.9 \\
        \hspace{0.8em}SIM-CoT & 55.6 & \textbf{55.9} & +0.3 & 61.1 & \textbf{62.2} & +1.1 & 97.2 & \textbf{97.8} & +0.6 & 21.1 \\
        \bottomrule
        \end{tabular}
    \vspace{-2pt}
\end{table}

\begin{table}[htbp]
    \centering
    \setlength{\abovecaptionskip}{2pt}
    \setlength{\belowcaptionskip}{3pt}
    \setlength{\aboverulesep}{0.3ex}
    \setlength{\belowrulesep}{0.3ex}
    \caption{\textbf{Comparison of latent CoT performance under adaptive and fixed-preset inference.} We report task accuracy (\%) under maximum latent budgets of 4, 5, and 6 steps.}
    \label{tab:frontier}
    \small
        \begin{tabular}{@{}l@{\hspace{6pt}}l@{\hspace{6pt}}c@{\hspace{4pt}}c@{\hspace{4pt}}c@{\hspace{8pt}}c@{\hspace{4pt}}c@{\hspace{4pt}}c@{\hspace{8pt}}c@{\hspace{4pt}}c@{\hspace{4pt}}c@{}}
        \toprule
        \multicolumn{2}{@{}l}{Model / Budget policy} & \multicolumn{9}{c}{Accuracy (\%)} \\
        \cmidrule(l){3-11}
        & & \multicolumn{3}{c}{GSM8K} & \multicolumn{3}{c}{SVAMP} & \multicolumn{3}{c}{MultiArith} \\
        \cmidrule(lr){3-5}\cmidrule(lr){6-8}\cmidrule(l){9-11}
        \multicolumn{2}{@{}l}{\emph{(a) CODI family}} & 4 steps & 5 steps & 6 steps & 4 steps & 5 steps & 6 steps & 4 steps & 5 steps & 6 steps \\
        \midrule
        \multicolumn{11}{@{}l}{\emph{GPT-2}} \\
        \hspace{0.8em}CODI & Fixed depth & 36.6 & 41.2 & 42.1 & 41.5 & 41.0 & 41.0 & 92.8 & 93.3 & 94.4 \\
         & \textbf{Adaptive} & \textbf{41.5} & \textbf{43.2} & \textbf{43.4} & \textbf{42.2} & \textbf{42.2} & \textbf{43.1} & 92.8 & 93.3 & \textbf{95.0} \\
        \addlinespace[2pt]
        \hspace{0.8em}SIM-CoT & Fixed depth & 35.3 & 39.1 & 39.2 & 41.3 & 40.6 & 40.0 & 89.4 & 91.1 & 91.7 \\
         & \textbf{Adaptive} & \textbf{40.6} & \textbf{41.0} & \textbf{41.5} & \textbf{42.8} & \textbf{42.6} & \textbf{43.5} & 89.4 & 91.1 & \textbf{92.2} \\
        \cmidrule(r){1-5}\cmidrule(lr){6-8}\cmidrule(l){9-11}
        \multicolumn{11}{@{}l}{\emph{Llama-1B}} \\
        \hspace{0.8em}CODI & Fixed depth & 48.0 & 54.7 & 55.5 & 61.1 & 61.6 & 61.1 & 97.2 & \textbf{97.8} & 97.2 \\
         & \textbf{Adaptive} & \textbf{53.8} & \textbf{55.0} & \textbf{55.7} & \textbf{62.0} & \textbf{61.7} & \textbf{62.3} & 97.2 & 97.2 & \textbf{98.3} \\
        \addlinespace[2pt]
        \hspace{0.8em}SIM-CoT & Fixed depth & 51.1 & 54.6 & 55.6 & 61.5 & 61.1 & 61.1 & \textbf{97.8} & 96.7 & 97.2 \\
         & \textbf{Adaptive} & \textbf{54.1} & \textbf{55.2} & \textbf{55.9} & \textbf{62.0} & \textbf{62.2} & \textbf{62.2} & 97.2 & \textbf{97.2} & \textbf{97.8} \\
        \midrule
        \multicolumn{2}{@{}l}{\emph{(b) Coconut family}} & 4 steps & 5 steps & 6 steps & 4 steps & 5 steps & 6 steps & 4 steps & 5 steps & 6 steps \\
        \midrule
        \multicolumn{11}{@{}l}{\emph{GPT-2}} \\
        \hspace{0.8em}Coconut & Fixed depth & 21.0 & \textbf{25.8} & 27.1 & 30.5 & 30.5 & 31.4 & 41.7 & 70.0 & 71.7 \\
         & \textbf{Adaptive} & 21.0 & 25.6 & \textbf{27.7} & \textbf{33.2} & \textbf{33.0} & \textbf{33.7} & \textbf{65.0} & \textbf{75.6} & \textbf{78.9} \\
        \addlinespace[2pt]
        \hspace{0.8em}SIM-CoT Coconut & Fixed depth & \textbf{27.9} & 34.6 & 38.2 & 40.1 & 40.0 & 40.7 & 92.8 & \textbf{92.2} & \textbf{92.8} \\
         & \textbf{Adaptive} & 26.5 & 34.6 & \textbf{40.4} & 40.1 & \textbf{40.7} & \textbf{41.6} & 92.8 & 91.1 & 91.1 \\
        \bottomrule
        \end{tabular}
    \vspace{-2pt}
\end{table}

\subsection{Constructive Progress Across Latent CoT}
Building on the advantages of adaptive latent CoT, we now examine constructive progress in trained latent reasoners. To understand why completing the full latent horizon can degrade reasoning performance, we first characterize how constructive progress evolves across latent depths, and then relate the progress to the reasoning performance.

\begin{figure}[htbp]
  \centering
    \setlength{\abovecaptionskip}  {3pt}
    \setlength{\belowcaptionskip}{0pt}
  \includegraphics[width=1.0\textwidth]{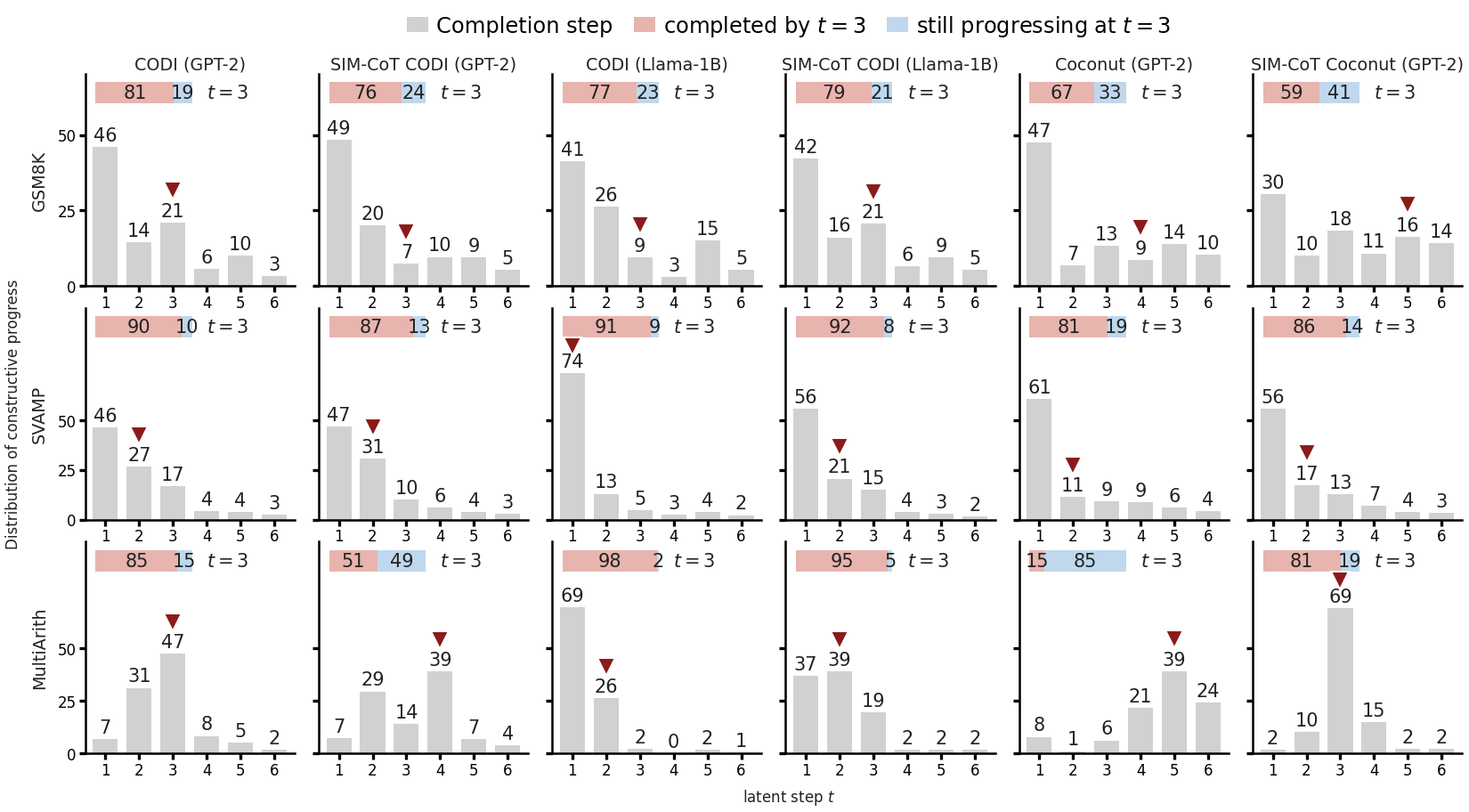}
    \caption{ \textbf{Constructive progress completion across latent CoT}. Bars show how constructive progress completion is distributed. The top strips summarize completion status at latent step $t=3$, contrasting completed and still progressing. Triangles mark the first latent step where at least $75\%$ of examples have completed constructive progress. Completion marks the first latent step where the latent CoT realizes all constructive progress.}
  \label{fig:sec5_2}
\end{figure}

Specifically, we characterize the distribution of constructive progress completion across latent steps. As shown in Figure~\ref{fig:sec5_2}, constructive progress is often established early, with a majority already completed by the third latent step across all six reasoners. Despite this common tendency, its progression differs across reasoners, with the CODI variants showing earlier completion while the Coconut variants retain more progress for later latent steps. These results indicate that trained latent reasoners can exhibit distinct constructive progress evolution. 

\begin{table}[htbp]
\centering
\setlength{\abovecaptionskip}{2pt}
\setlength{\belowcaptionskip}{3pt}
\setlength{\aboverulesep}{0.3ex}
\setlength{\belowrulesep}{0.3ex}
\setlength{\tabcolsep}{4pt}
\caption{\textbf{Reasoning performance across constructive progress.} At latent depth $t = 3$, we compare task accuracy (\%) across different levels of realized constructive progress, as defined in Section~\ref{sec:4.2}.}
\label{tab:progress_performance_nobold}
\small
\begin{tabular*}{\textwidth}{@{\extracolsep{\fill}}llcccccc@{}}
\toprule
\multirow{2}{*}{Backbone} & \multirow{2}{*}{Method} & \multicolumn{2}{c}{GSM8K} & \multicolumn{2}{c}{SVAMP} & \multicolumn{2}{c}{MultiArith} \\
\cmidrule(lr){3-4}\cmidrule(lr){5-6}\cmidrule(l){7-8}
& & Lower & Higher & Lower & Higher & Lower & Higher \\
\midrule
\emph{GPT-2} & CODI & 29.3 & 45.0 & 16.6 & 70.8 & 81.0 & 96.7 \\
& SIM-CoT & 10.8 & 53.3 & 13.2 & 73.4 & 25.3 & 94.1 \\
\cmidrule{1-8}
\emph{Llama-1B} & CODI & 24.9 & 69.8 & 54.4 & 77.0 & 95.1 & 99.0 \\
& SIM-CoT & 34.7 & 67.2 & 45.3 & 80.3 & 94.5 & 98.4 \\
\midrule
\emph{GPT-2} & Coconut & 11.4 & 32.8 & 21.1 & 45.3 & 31.4 & 38.7 \\
& SIM-CoT Coconut & 6.9 & 47.2 & 26.5 & 66.9 & 77.1 & 94.7 \\
\bottomrule
\end{tabular*}
\end{table}

We further examine the relation between realized constructive progress and reasoning performance. At the third latent step, we compare task accuracy across different levels of realized constructive progress. As shown in Table~\ref{tab:progress_performance_nobold}, higher constructive progress consistently corresponds to higher reasoning accuracy. For instance, on SVAMP with GPT-2/SIM-CoT, accuracy increases from \(13.2\%\) under Lower progress to \(73.4\%\) under Higher progress. These results show that reasoning performance mainly depends on how much constructive progress has already been realized.

These results indicate that the same latent depth can correspond to different levels of constructive progress. At a fixed latent depth, reasoning performance varies systematically with the realized constructive progress. This relation is also consistent with the gains from adaptive inference in Table~\ref{tab:results_main}. Adaptive inference can adjust the executed horizon of CoT according to the evolution of constructive progress, while a fixed latent CoT cannot.

\subsection{Remaining Progress and Computation Returns}
\label{sec:5.3}
So far, we have shown that latent reasoning outcomes vary with constructive progress along the reasoning frontier. Here, we test the relation between remaining progress and future gains from continued latent CoT. Recall from Section~\ref{sec:4.2} that $\Omega_3$ denotes the remaining constructive progress at the third latent step. We compare task accuracy at Steps~3 and~4 under lower and higher $\Omega_3$. As shown in Table~\ref{tab:5.3}(a), task accuracy consistently improves more from the third to the fourth latent step under higher $\Omega_3$. Interestingly, higher $\Omega_3$ can coincide with lower accuracy at the third step while yielding substantially larger gains after continuation. For instance, with GPT-2/SIM-CoT, accuracy remains at $35.0\%$ under lower $\Omega_3$, whereas it increases from $18.5\%$ to $57.3\%$ under higher $\Omega_3$. Therefore, we argue that more remaining constructive progress corresponds to a larger reasoning gain from continuing latent CoT.

We further extend this comparison over the subsequent latent steps, using Step 3 as the common starting point. As shown in Table~\ref{tab:5.3}(b), the future gain difference associated with higher $\Omega_3$ grows over subsequent latent steps. Notably, this difference can remain small after one latent step but become larger as latent CoT continues. For instance, with Llama-1B/CODI, the difference is only $1.6\%$ at Step~4, but later expands to $58.4\%$. These results indicate that the future gain indicated by remaining constructive progress can be realized cumulatively over subsequent latent CoT.

\begin{table}[htbp]
\centering
\footnotesize
\setlength{\tabcolsep}{3pt}
\setlength{\gainwd}{15pt}
\setlength{\abovecaptionskip}{2pt}
\setlength{\belowcaptionskip}{3pt}
\sisetup{retain-explicit-plus}
\def\pgap{\hspace{10pt}}
\caption{\textbf{Reasoning gains across remaining constructive progress.}
$\Omega_3$ means the remaining constructive progress after 3 latent steps.
Panel \textbf{(a)} reports task accuracy at Steps 3 and 4 under lower and higher $\Omega_3$. Panel \textbf{(b)} reports future reasoning gain, measured by the change in $\hat{G}_t$ from Step 3 to Steps 4 and 5, under lower and higher $\Omega_3$, with subscripts indicating the difference relative to lower $\Omega_3$. Panel \textbf{(c)} reports the change in future reasoning gain as $\Omega_t$ decreases, with negative values indicating a decrease. We report GSM8K here and defer additional results to Appendix~\ref{app:Additional Results Remaining Progress}.}
\label{tab:5.3}
\begin{adjustbox}{max width=\linewidth}
\begin{tabular}{@{}ll@{\pgap}S[table-format=2.1]S[table-format=2.1]@{}lS[table-format=2.1]S[table-format=2.1]@{}l@{\pgap}S[table-format=+1.1]S[table-format=+1.1]S[table-format=+2.1]@{}lS[table-format=+2.1]@{}l@{\pgap}S[table-format=+2.1]S[table-format=+2.1]S[table-format=+2.1]@{}}
\toprule
&&\multicolumn{6}{c@{\pgap}}{(a) Accuracy (\%)}&\multicolumn{6}{c@{\pgap}}{(b) Future gain}&\multicolumn{3}{c@{}}{(c) Future gain change}\\
\cmidrule(l{2pt}r{12pt}){3-8}\cmidrule(l{2pt}r{12pt}){9-14}\cmidrule(l{2pt}r{0pt}){15-17}
&&\multicolumn{3}{c}{Lower $\Omega_3$}&\multicolumn{3}{c@{\pgap}}{Higher $\Omega_3$}&\multicolumn{2}{c}{Lower $\Omega_3$}&\multicolumn{4}{c@{\pgap}}{Higher $\Omega_3$}&\multicolumn{3}{c@{}}{Decrease in $\Omega_t$}\\
\cmidrule(l{2pt}r{2pt}){3-5}\cmidrule(l{2pt}r{12pt}){6-8}\cmidrule(l{2pt}r{2pt}){9-10}\cmidrule(l{2pt}r{12pt}){11-14}\cmidrule(l{2pt}r{0pt}){15-17}
\multicolumn{2}{@{}l}{Backbone / Method}&{Step 3}&{Step 4}&&{Step 3}&{Step 4}&&{Step 4}&{Step 5}&{Step 4}&&{Step 5}&&{Lower}&{Middle}&{Higher}\\
\midrule
\multicolumn{17}{@{}l}{\textit{CODI family}}\\
\textit{GPT-2}&CODI&46.8&46.3&\gdn{0.5}&34.0&35.9&\gup{1.9}&-0.2&-0.2&+4.9&\gup{5.1}&+44.0&\gup{44.2}&+0.5&-3.7&-49.9\\
&SIM-CoT&35.0&35.0&\gz&18.5&57.3&\gup{38.8}&+0.1&+0.2&+37.0&\gup{36.9}&+71.8&\gup{71.6}&+0.4&-3.3&-47.5\\
\addlinespace[3pt]
\textit{Llama-1B}&CODI&39.5&39.5&\gz&23.9&31.0&\gup{7.1}&-0.1&+1.4&+1.5&\gup{1.6}&+59.8&\gup{58.4}&-1.2&-9.7&-47.7\\
&SIM-CoT&51.6&50.5&\gdn{1.1}&23.7&56.3&\gup{32.6}&-1.5&+2.0&+25.7&\gup{27.2}&+53.1&\gup{51.1}&-3.5&-10.7&-53.6\\
\midrule
\multicolumn{17}{@{}l}{\textit{Coconut family}}\\
\textit{GPT-2}&Coconut&16.8&16.4&\gdn{0.4}&33.0&47.0&\gup{14.0}&+0.2&+0.3&+11.7&\gup{11.5}&+41.2&\gup{40.9}&+1.7&+1.8&-28.3\\
&SIM-CoT Coconut&27.8&30.8&\gup{3.0}&2.3&41.7&\gup{39.4}&+3.0&+2.6&+33.8&\gup{30.8}&+79.7&\gup{77.1}&+5.9&-6.0&-66.4\\
\bottomrule
\end{tabular}
\end{adjustbox}
\end{table}

As these constructive gains are realized, however, the remaining progress changes in turn. Further frontier advancement therefore leaves less further constructive progress. We therefore examine how the future gain changes as $\Omega_t$ decreases along this progression. As shown in Table~\ref{tab:5.3}(c), larger decreases in $\Omega_t$ are associated with larger reductions in the subsequent reasoning gain. The reduction becomes more pronounced as more constructive progress is realized. These results show that the future gain from continued latent CoT diminishes as constructive progress advances the reasoning frontier. Consequently, these results show that future gains from latent CoT evolve systematically with constructive progress along the reasoning frontier.
\subsection{Continuation Signals for Adaptive Latent Inference}
\label{sec:signals}
 
We have shown that remaining progress can guide adaptive latent inference on frozen reasoners. To provide some insights into adaptive latent inference, we further compare remaining progress with two groups of continuation signals. We first consider online signals that we derive from the trajectory representation $X_t$, including answer stability, prediction confidence, and state convergence (Table~\ref{tab:continuation_signals}). We then try to transfer the core concepts of some adaptive methods, which mostly learn when to stop during training, into continuation signals (Table~\ref{tab:rebuttal_rules}). Notably, these concepts mainly characterize the current state or the input. Specifically, ALR~\citep{ning2025learning} decides whether to stop from the current latent state, DEER~\citep{yang2026deer} exits once the confidence of the current prediction is high, and Dynamic K-Head~\citep{li2026dynamickhead} predicts a latent budget from the question before latent reasoning starts.
 
As shown in Table~\ref{tab:continuation_signals}, stability and confidence perform well in several settings, indicating that the latent trajectory carries information about future gains. In Table~\ref{tab:rebuttal_rules}, remaining progress yields higher accuracy than all transferred criteria across settings, by up to 4.9 percentage points. Notably, the gap is largest on GSM8K, where continued latent computation still yields gains. For instance, with GPT-2/SIM-CoT on GSM8K, accuracy is 39.2\% under remaining progress, compared with 35.1\%, 34.3\%, and 34.3\% under ALR, DEER, and Dynamic K-Head. Replacing our estimate with the true remaining progress further improves accuracy in every setting.
 
These results are expected, since adapting the latent horizon requires accounting for how the reasoning benefits of latent CoT evolve. Criteria based on the current state or the input decide from what has already been computed, whereas remaining progress anticipates what continued latent computation can still achieve. Therefore, we propose a constructive-progress framework to draw the potential gains of continued latent CoT, and use remaining progress as the online signal throughout this paper.

\begin{table}[htbp]
\centering
\footnotesize
\setlength{\tabcolsep}{3pt}
\setlength{\gainwd}{15pt}
\setlength{\abovecaptionskip}{2pt}
\setlength{\belowcaptionskip}{3pt}
\sisetup{retain-explicit-plus}
\def\pgap{\hspace{10pt}}
\caption{\textbf{Continuation signals under matched latent computation.} All criteria use the same adaptive controller and differ only in the online signal used to allocate the remaining latent budget. We report final task accuracy under matched average latent computation, with subscripts indicating the difference relative to remaining progress.}
\label{tab:continuation_signals}
\begin{adjustbox}{max width=\linewidth}
\begin{tabular}{@{}lll@{\pgap}S[table-format=2.1]S[table-format=2.1]@{}lS[table-format=2.1]@{}lS[table-format=2.1]@{}l@{\pgap}S[table-format=1.2]@{}}
\toprule
&&&\multicolumn{7}{c@{\pgap}}{Accuracy (\%) by continuation signal}&\multicolumn{1}{c@{}}{Compute}\\
\cmidrule(l{2pt}r{12pt}){4-10}\cmidrule(l{2pt}r{0pt}){11-11}
Dataset&Backbone&Method&{Progress}&\multicolumn{2}{c}{Stability}&\multicolumn{2}{c}{Confidence}&\multicolumn{2}{c@{\pgap}}{Convergence}&\multicolumn{1}{c@{}}{Avg.\ steps}\\
\midrule
GSM8K&\textit{GPT-2}&CODI&41.1&40.5&\gdn{0.6}&40.2&\gdn{0.9}&40.0&\gdn{1.1}&4.37\\
&&SIM-CoT&39.2&37.9&\gdn{1.3}&36.8&\gdn{2.4}&37.5&\gdn{1.7}&4.28\\
\addlinespace[2pt]
&\textit{Llama-1B}&SIM-CoT&54.7&53.3&\gdn{1.4}&54.2&\gdn{0.5}&53.4&\gdn{1.3}&4.72\\
\midrule
SVAMP&\textit{GPT-2}&SIM-CoT&40.2&40.7&\gup{0.5}&40.7&\gup{0.5}&40.1&\gdn{0.1}&4.68\\
&\textit{Llama-1B}&CODI&60.7&60.9&\gup{0.2}&61.0&\gup{0.3}&60.5&\gdn{0.2}&4.85\\
\midrule
MultiArith&\textit{GPT-2}&SIM-CoT&89.4&88.9&\gdn{0.5}&91.7&\gup{2.3}&88.3&\gdn{1.1}&4.61\\
\bottomrule
\end{tabular}
\end{adjustbox}
\end{table}

\begin{table}[htbp]
\centering
\setlength{\abovecaptionskip}{2pt}
\setlength{\belowcaptionskip}{3pt}
\setlength{\aboverulesep}{0.3ex}
\setlength{\belowrulesep}{0.3ex}
\setlength{\tabcolsep}{3.5pt}
\setlength{\gainwd}{15pt}
\caption{\textbf{Comparison with transferred adaptive latent methods.} All methods run on the same frozen reasoners with matched average latent computation, and none stops before our adaptation step. The upper bounds replace our estimate with the true remaining progress, or give each item its best stopping step under the same total computation. Subscripts indicate the difference relative to ours.}
\label{tab:rebuttal_rules}
\small
\begin{tabular}{@{}lllcccc@{\hspace{8pt}}cc@{\hspace{8pt}}c@{}}
\toprule
 & & & \multicolumn{4}{c}{Accuracy (\%)} & \multicolumn{2}{c}{Upper bound} & Compute \\
\cmidrule(lr){4-7}\cmidrule(lr){8-9}\cmidrule(l){10-10}
Dataset & Backbone & Method & Ours & ALR & DEER & \makecell{Dynamic\\K-Head} & \makecell{Ours with\\true $\Omega$} & \makecell{Per-item\\oracle} & Avg. steps \\
\midrule
GSM8K & \emph{GPT-2} & CODI & 41.1 & 38.4\gdn{2.7} & 37.3\gdn{3.8} & 38.4\gdn{2.7} & 42.6 & 47.0 & 4.37 \\
& & SIM-CoT & 39.2 & 35.1\gdn{4.1} & 34.3\gdn{4.9} & 34.3\gdn{4.9} & 41.8 & 44.7 & 4.28 \\
& \emph{Llama-1B} & CODI & 54.5 & 53.1\gdn{1.4} & 54.4\gdn{0.1} & 52.5\gdn{2.0} & 55.6 & 58.8 & 4.57 \\
& & SIM-CoT & 54.7 & 52.3\gdn{2.4} & 52.8\gdn{1.9} & 52.9\gdn{1.8} & 55.5 & 58.2 & 4.72 \\
\midrule
SVAMP & \emph{GPT-2} & SIM-CoT & 40.2 & 39.9\gdn{0.3} & 39.6\gdn{0.6} & 39.8\gdn{0.4} & 41.5 & 42.9 & 4.68 \\
& \emph{Llama-1B} & CODI & 60.7 & 60.4\gdn{0.3} & 60.3\gdn{0.4} & 60.5\gdn{0.2} & 61.2 & 61.6 & 4.85 \\
\midrule
MultiArith & \emph{GPT-2} & SIM-CoT & 89.4 & 88.9\gdn{0.5} & 87.1\gdn{2.3} & 87.2\gdn{2.2} & 91.7 & 92.8 & 4.61 \\
& \emph{Llama-1B} & CODI & 97.2 & 96.6\gdn{0.6} & 96.9\gdn{0.3} & 96.6\gdn{0.6} & 97.8 & 98.3 & 4.59 \\
\bottomrule
\end{tabular}
\end{table}

\section{Conclusion}

The intuition behind completing a preset latent horizon is that additional latent steps should provide more opportunity for reasoning. However, our study shows that reasoning gains do not simply accumulate with latent depth. Completing the full horizon can overwrite previously reached gold answers and diminish existing reasoning gains. More broadly, our results show that higher realized constructive progress corresponds to higher current reasoning accuracy, while more remaining progress corresponds to larger future gains from continued latent CoT. By rethinking the fixed-horizon of latent CoT, we develop a method that adaptively adjusts the executed horizon during inference. Across widely studied latent CoT approaches, including CODI, Coconut, and SIM-CoT variants, and multiple arithmetic benchmarks, our method consistently improves reasoning accuracy and also reduces latent computation.

\subsection*{AI use statement}
In this work, we did not use generative AI tools for tasks requiring mandatory disclosure under the ICLR 2027 AI Policy. Additionally, we used generative AI tools to create or edit software code and to edit the research paper to improve readability. We have reviewed all AI-assisted work. AI-assisted code was verified and tested by the authors, and AI-assisted edits to the manuscript were reviewed by the authors. We take responsibility for the final content of this work, including text, claims or artifacts produced with the aid of generative AI.

\subsection*{Reproducibility statement}
The source code used for our experiments is provided in the supplementary materials and will be released in a public repository. Appendix B provides the experimental details for task-level evaluation, adaptive latent inference, and latent progression analysis. Appendix C provides the assumptions and complete proofs for the theoretical results. Appendix D reports additional experimental results.

\bibliography{iclr2027_conference}
\bibliographystyle{iclr2027_conference}

\clearpage
\appendix

\titleformat{\section}
  {\large\bfseries}
  {\thesection.}
  {0.6em}
  {}

\titleformat{\subsection}
  {\normalsize\bfseries}
  {\thesubsection.}
  {0.6em}
  {}

\titlespacing*{\section}
  {0pt}
  {2.0ex plus 0.4ex minus 0.2ex}
  {1.0ex}

\titlespacing*{\subsection}
  {0pt}
  {1.7ex plus 0.4ex minus 0.2ex}
  {0.8ex}

\setlength{\abovedisplayskip}{6pt plus 2pt minus 2pt}
\setlength{\belowdisplayskip}{6pt plus 2pt minus 2pt}
\setlength{\abovedisplayshortskip}{4pt plus 1pt minus 1pt}
\setlength{\belowdisplayshortskip}{4pt plus 1pt minus 1pt}

\noindent{\Large\bfseries Appendix}
\vspace{1.5ex}

\section{Related Work}
\label{app:related-work}

Recent work has developed several formulations of latent CoT, ranging from implicit reasoning through distillation~\citep{deng2023icot} to direct reasoning in continuous latent space~\citep{hao2025coconut}. Meanwhile, latent CoT has also been studied through self-distillation~\citep{shen2025codi}, soft representations~\citep{xu2025softcot}, and supervised implicit CoT~\citep{wei2026simcot}. These studies establish latent CoT as a broad family of reasoning methods acquired through distinct training schemes. One common structure is that intermediate reasoning unfolds through successive latent states before explicit answer generation. Such latent trajectories have been considered a common structure for multi-step inference across different latent CoT methods~\citep{chen2025latentcotsurvey}.

While early work primarily focused on training latent reasoners, recent studies have increasingly examined the mechanisms underlying latent CoT. Specifically, at the representation level, some works characterize latent states as compressed representations of explicit reasoning~\citep{dilgren2026interpretable}. Meanwhile, some studies show that continuous reasoning can preserve multiple reasoning alternatives in superposition~\citep{zhu2025superposition}, enabling parallel exploration over candidate reasoning traces~\citep{gozeten2026cot2}. More recently, the community has also begun to examine how latent reasoning evolves across successive steps, both through its training dynamics~\citep{zhu2026emergence} and through probing of internal representations~\citep{rizvimartel2026illusion}. Taken together, existing work has predominantly focused on characterizing the structure and evolution of latent reasoning, leaving less examined how its reasoning benefits evolve along the latent trajectory. Our work complements these studies by characterizing how the reasoning value of latent CoT changes along this progression.

The core insight of adaptive reasoning methods is to vary the reasoning depth across different instances~\citep{graves2016act,dehghani2019universal}. Early approaches often set the reasoning depth according to problem difficulty~\citep{banino2021pondernet,snell2025ttc}, while recent methods increasingly adapt it using confidence or consistency in intermediate predictions~\citep{aggarwal2023adaptive,yang2026deer}. Yet, these criteria primarily characterize problem difficulty or the current prediction, leaving the value of further reasoning less directly examined. Dynamic compression has further extended this adaptive view to latent reasoning chains~\citep{tan2025colar}. In latent reasoning, recurrent-depth models exit once successive latent steps converge~\citep{geiping2026scaling}, while recent methods learn when to stop through reinforcement learning~\citep{ning2025learning} or predict a latent budget from the question~\citep{li2026dynamickhead}. These criteria mainly aim to save computation, without considering where the gains from latent reasoning come from. Nevertheless, these methods generally assume that additional latent steps can yield further reasoning gain. In this paper, we revisit this assumption, characterize where latent CoT gains arise, and develop a method that adapts its depth accordingly to improve reasoning accuracy.

\section{Experimental Details}
\label{app:setup}
\subsection{Task-Level Evaluation}
\label{app:Experimental Details-1}

\textbf{Latent depth evaluation.} For task-level evaluation, we decode a complete answer from each evaluated latent prefix. Within each reasoner, intermediate depths and Full-$K$ use the same answer-decoding procedure, differing only in the latent prefix length.

\textbf{Benchmarks and scoring.} For arithmetic reasoning tasks, we evaluate on the official GSM8K test split, the full SVAMP benchmark, and the MultiArith test split. We use greedy decoding throughout and follow the original evaluation protocol of each latent reasoner family for final-answer extraction and matching. We instantiate $r(x,y)$ as binary answer correctness. Each output receives $r(x,y)=1$ if the extracted answer matches the ground-truth answer, and $0$ otherwise. We report accuracy as the mean binary correctness over the full evaluation set, with outputs lacking a valid extracted answer counted as incorrect.

\textbf{Horizon configuration.} Throughout our evaluation, we use a continuous thought, or latent state, as the basic unit of latent depth. For CODI and SIM-CoT CODI, one continuous thought corresponds to one latent step. For Coconut and SIM-CoT Coconut, each latent stage contains two continuous thoughts, corresponding to two latent steps. The native preset horizon is six latent steps for CODI, SIM-CoT CODI, and Coconut, and ten latent steps (five stages) for SIM-CoT Coconut. We use Full-$K$ to denote the native preset horizon of each latent reasoner. For analyses that require a shared depth axis, we use the first six latent steps as a common analysis window.

\subsection{Adaptive Latent Inference}
\label{app:Experimental Details-2}
\paragraph{Trajectory representation.}
For adaptive latent inference, $X_t$ summarizes the observed evolution of the latent trajectory up to step $t$. We construct $X_t$ from predictive statistics, the latent state, and their short-range changes over the observed prefix. The predictive statistics capture confidence, distributional change, and prediction stability, while the latent state is projected to a 64-dimensional representation together with its first-order change. We use observations from the current and two preceding latent steps, with unavailable history zero-filled together with an explicit missing indicator. The PCA projection is fitted on offline training trajectories.

\paragraph{Remaining-progress estimation.}
We construct the remaining-progress targets from offline trajectories of the fixed latent reasoner. At each latent depth, we estimate $\hat{G}_t$ as the fraction of sampled continuations that produce the correct answer. We then construct $\hat{\Omega}_t$ from these reasoning-value estimates according to Eq.~\eqref{eq:remaining_progress} and fit $f_\phi$ to predict $\hat{\Omega}_t$ from $X_t$ by squared-error regression. The estimator is fitted separately for each model--task configuration using its offline trajectories. We implement $f_\phi$ with a histogram-based gradient-boosting regressor.

\paragraph{Horizon adaptation.}
For each model configuration, we construct $F_{\mathrm{ref}}$ as the empirical cumulative distribution of reference progress estimates and keep it fixed when evaluating the current examples. With $u_t$ defined in Eq.~\eqref{eq:relative_progress}, we instantiate the continuation rule as
\[
g(u_t)
=
\operatorname{clip}
\left(
\rho_0 + 0.5(2u_t-1),\,0,\,1
\right),
\]
where $\rho_0$ specifies the baseline fraction of the remaining latent horizon. The additional latent budget is $a_t=(K-t)g(u_t)$, following Eq.~\eqref{eq:latent_allocation}. We apply the estimator and continuation rule once at the current adaptation step and use randomized rounding to convert the resulting budget into an integer realized horizon $\tau\leq K$. For a nonnegative allocation $a$, randomized rounding returns $\lfloor a\rfloor+1$ with probability $a-\lfloor a\rfloor$, and $\lfloor a\rfloor$ otherwise.

\paragraph{Continuation signals.}
For the comparison in Section~\ref{sec:signals}, we instantiate continuation signals from the information summarized by $X_t$. Answer stability counts the consecutive latent steps over which the decoded answer remains unchanged, prediction confidence is the log-likelihood of the decoded answer at the current step, and state convergence is the normalized change of the projected latent state between consecutive steps. Each signal is read at the adaptation step and mapped through $F_{\mathrm{ref}}$ as for the estimated remaining progress, with $\rho_0$ chosen to match the average latent computation.

\subsection{Latent Progression Analysis}
\label{app:Experimental Details-3}
For the analysis in ProsQA, each node represents a reasoning candidate, and the target node denotes the gold answer. At each latent step, the continuous thought induces a vocabulary distribution, from which we record the probability of each reasoning candidate. We use these probabilities together with graph depth to trace candidate progression over latent steps. Across the full ProsQA set, candidates at larger graph depths tend to receive support at later latent steps, with a Spearman correlation of $0.833$. After the gold answer becomes the most supported candidate, later latent steps continue to change the leading candidate, with correction and reversal rates of $16.1\%$ and $15.6\%$, respectively.

For the realization intervention in Figure~\ref{fig:3.2}, we replace the latent state $z_t$ with the embedding of its top token $v^\star$, and the matched control instead adds to $z_t$ a random unit direction scaled to $\lVert e(v^\star) - z_t \rVert$; both are applied once at the same latent step.

Across the six arithmetic reasoners, we characterize progression directly from task outcomes over latent depth. Constructive progress is realized at different latent depths across reasoners, occurring earlier for the CODI variants and later for the Coconut variants. Consistent with Figure~\ref{fig:1}, the correctness trajectories also show that outcomes reached at intermediate depths can change under continued latent computation.

\section{Proofs and Derivations}
\label{app:proofs}
\subsection{Population Estimation of Remaining Progress}
\label{app:causal_estimation}
For completeness, we restate Proposition~\ref{prop:causal-estimation}, which characterizes the optimal population prediction of remaining progress from $X_t$ under squared risk, in its precise form and prove it.
\begin{restated}{Proposition}{prop:causal-estimation}{Causal Estimation of Remaining Progress}
For the population squared risk
$\mathcal{R}(f)=\mathbb{E}[(f(X_t)-\Omega_t)^2]$
over square-integrable predictors $f(X_t)$, the minimizer satisfies
\[
f^*(X_t)=\mathbb{E}[\Omega_t\mid X_t]
\qquad \text{a.s.}
\]
\end{restated}

\begin{proof}
Fix a latent depth $t$ and write
$m_t(X_t):=\mathbb{E}[\Omega_t\mid X_t]$.
For any square-integrable predictor $f(X_t)$,
\begin{align*}
\mathcal{R}(f)
&=
\mathbb{E}\!\left[
\bigl(\Omega_t-m_t(X_t)+m_t(X_t)-f(X_t)\bigr)^2
\right] \\
&=
\mathbb{E}\!\left[
\bigl(\Omega_t-m_t(X_t)\bigr)^2
\right]
+
\mathbb{E}\!\left[
\bigl(m_t(X_t)-f(X_t)\bigr)^2
\right] \\
&\quad+
2\mathbb{E}\!\left[
\bigl(m_t(X_t)-f(X_t)\bigr)
\mathbb{E}\!\left[
\Omega_t-m_t(X_t)\mid X_t
\right]
\right].
\end{align*}
Since
$\mathbb{E}[\Omega_t-m_t(X_t)\mid X_t]=0$,
the cross term vanishes. Hence,
\begin{equation}
\mathcal{R}(f)
=
\mathcal{R}(m_t)
+
\mathbb{E}\!\left[
\bigl(f(X_t)-m_t(X_t)\bigr)^2
\right].
\label{eq:remaining_progress_risk_decomposition}
\end{equation}
The second term is nonnegative and vanishes if and only if
$f(X_t)=m_t(X_t)$ almost surely. Therefore,
$f^*(X_t)=m_t(X_t)=\mathbb{E}[\Omega_t\mid X_t]$ almost surely,
which proves Proposition~\ref{prop:causal-estimation}.
\end{proof}

At the optimum,
$\mathcal{R}(f^*)=
\mathbb{E}[\operatorname{Var}(\Omega_t\mid X_t)]$.
Equation~\eqref{eq:remaining_progress_risk_decomposition} therefore
separates the minimum prediction risk given $X_t$ from the excess risk
of a predictor relative to the conditional mean.
\subsection{Frontier Progress and Additional Reasoning Gains}
\label{app:frontier_gain}
We make explicit the relation between the remaining frontier progress
and the reasoning gain realized by additional latent computation.
Fix a latent depth $t$ and write
$M_t := \max_{1\leq j\leq t} G_j$, so that
$\Omega_t=M_K-M_t$.

\noindent\textbf{Frontier realization.}
For an integer $a\in\{0,\ldots,K-t\}$, define the frontier advancement
realized within $a$ additional latent steps as
\begin{equation}
P_t(a)
:=
M_{t+a}-M_t
=
\max_{1\leq j\leq t+a}G_j
-
\max_{1\leq j\leq t}G_j .
\label{eq:frontier_realization}
\end{equation}

\begin{applemma}[Frontier realization over the remaining horizon]
\label{lem:frontier-realization}
For any integers $0\leq a\leq b\leq K-t$,
\[
0=P_t(0)\leq P_t(a)\leq P_t(b)\leq\Omega_t,
\qquad
P_t(K-t)=\Omega_t .
\]
\end{applemma}

\begin{proof}
The index sets are nested as
\[
\{1,\ldots,t\}
\subseteq
\{1,\ldots,t+a\}
\subseteq
\{1,\ldots,t+b\}
\subseteq
\{1,\ldots,K\}.
\]
Hence, the corresponding frontier values are nondecreasing, which gives
$0=P_t(0)\leq P_t(a)\leq P_t(b)\leq\Omega_t$.
Finally, setting $a=K-t$ in
Eq.~\eqref{eq:frontier_realization} gives
\[
P_t(K-t)
=
M_K-M_t
=
\Omega_t,
\]
which proves the result.
\end{proof}

\noindent\textbf{Signed reasoning gain.}
Frontier advancement does not by itself determine whether the reasoning
value after continuation improves over its current value. We therefore
define the signed reasoning gain from $a$ additional latent steps as
$\Delta G_t(a):=G_{t+a}-G_t$.
Let $d_t:=M_t-G_t\geq 0$ denote the gap between the current reasoning
value and the best frontier reached so far.

\begin{applemma}[Decomposition of additional reasoning gain]
\label{lem:gain-decomposition}
For any $a\in\{0,\ldots,K-t\}$,
\begin{equation}
\Delta G_t(a)
=
P_t(a)+d_t-d_{t+a}.
\label{eq:gain_decomposition}
\end{equation}
\end{applemma}

\begin{proof}
By definition,
$G_t=M_t-d_t$ and $G_{t+a}=M_{t+a}-d_{t+a}$.
Therefore,
\[
\begin{aligned}
\Delta G_t(a)
&=
G_{t+a}-G_t \\
&=
\bigl(M_{t+a}-d_{t+a}\bigr)
-
\bigl(M_t-d_t\bigr) \\
&=
P_t(a)+d_t-d_{t+a},
\end{aligned}
\]
which proves Eq.~\eqref{eq:gain_decomposition}.
\end{proof}

In particular, over the full remaining horizon,
\[
\Delta G_t(K-t)
=
\Omega_t+d_t-d_K.
\]
Thus, remaining frontier progress and the endpoint gain need not
coincide: whether the final reasoning value improves also depends on how
the trajectory moves relative to the frontier already attained.

Moreover, if $\Omega_t=0$, Lemma~\ref{lem:frontier-realization} gives $P_t(a)=0$ for every
$a\in\{0,\ldots,K-t\}$, and Eq.~\eqref{eq:gain_decomposition} reduces to
\[
\Delta G_t(a)=d_t-d_{t+a}.
\]
In this case, additional computation cannot establish a new frontier;
its signed gain is determined entirely by how the subsequent state
moves relative to the existing frontier.

\noindent\textbf{Expected continuation return.}
For $\omega$ in the support of $\Omega_t$, define
\begin{equation}
V_t(a,\omega)
:=
\mathbb{E}\!\left[
\Delta G_t(a)\mid\Omega_t=\omega
\right].
\label{eq:expected_continuation_return}
\end{equation}
At a fixed depth $t$, we suppress the subscript and write
$V(a,\omega)$. By construction, $V_t(0,\omega)=0$.
Unlike the frontier realization $P_t(a)$, both
$\Delta G_t(a)$ and $V_t(a,\omega)$ are signed quantities and therefore
capture whether additional latent computation improves or degrades the
reasoning value.

\subsection{Monotone Allocation by Remaining Progress}
\label{app:monotone_allocation}
This subsection proves Theorem~\ref{thm:monotone-allocation}, which characterizes how optimal additional computation varies with remaining progress.

\noindent\textbf{Discrete allocation.}
Fix a latent depth $t$ and let $A:=K-t$. For trajectories
$i=1,\ldots,n$ with remaining progress $\omega_i$, define the feasible
allocation set
\[
\mathcal{A}_B
:=
\left\{
a\in\{0,\ldots,A\}^n:
\sum_{i=1}^{n} a_i\leq B
\right\}.
\]
Assume that $V(a,\omega)$ has increasing differences in $(a,\omega)$.
Thus, for integers $0\leq a\leq a'\leq A$ and $\omega\leq\omega'$,
\[
V(a',\omega')-V(a,\omega')
\geq
V(a',\omega)-V(a,\omega).
\]

\begin{proof}[Proof of Theorem~\ref{thm:monotone-allocation}]
Since $\mathcal{A}_B$ is finite and nonempty, an optimal allocation
exists. Let $\bar a\in\mathcal{A}_B$ be an optimal allocation. Relabel the trajectories so that $\omega_1\geq\cdots\geq\omega_n$. If there exist $i<j$ with $\omega_i>\omega_j$ but $\bar a_i<\bar a_j$, exchange $\bar a_i$ and $\bar a_j$. The exchange preserves feasibility. By increasing differences,
\[
V(\bar a_j,\omega_i)+V(\bar a_i,\omega_j)
\geq
V(\bar a_i,\omega_i)+V(\bar a_j,\omega_j).
\]
Hence, removing such an inversion does not decrease the objective.
Applying these exchanges until no inversion remains yields an optimal
allocation $a^*$ satisfying
\[
\omega_i>\omega_j
\quad\Longrightarrow\quad
a_i^*\geq a_j^*,
\]
which proves Theorem~\ref{thm:monotone-allocation}.
\end{proof}

\subsection{Marginal Returns and Remaining Progress}
\label{app:marginal_returns_C}
We now make explicit the marginal-return interpretation of the
increasing-differences condition in Theorem~\ref{thm:monotone-allocation}. Fix a latent depth $t$.
For integers $0\leq a<a'\leq K-t$, define the incremental return from
extending the additional latent budget from $a$ to $a'$ as
\[
R_t(a,a')
:=
\Delta G_t(a')-\Delta G_t(a)
=
G_{t+a'}-G_{t+a}.
\]
\begin{applemma}[Decomposition of incremental continuation return]
\label{lem:incremental-return}
For any integers $0\leq a<a'\leq K-t$,
\begin{equation}
R_t(a,a')
=
P_t(a')-P_t(a)
+
d_{t+a}-d_{t+a'} .
\label{eq:incremental_return_decomposition}
\end{equation}
Moreover, for $\omega$ in the support of $\Omega_t$,
\begin{equation}
V_t(a',\omega)-V_t(a,\omega)
=
\mathbb{E}\!\left[
R_t(a,a')
\mid
\Omega_t=\omega
\right].
\label{eq:population_incremental_return}
\end{equation}
\end{applemma}

\begin{proof}
Subtracting the decomposition in
Eq.~\eqref{eq:gain_decomposition} at budgets $a$ and $a'$ gives
\[
\begin{aligned}
R_t(a,a')
&=
\Delta G_t(a')-\Delta G_t(a) \\
&=
\bigl(P_t(a')+d_t-d_{t+a'}\bigr)
-
\bigl(P_t(a)+d_t-d_{t+a}\bigr) \\
&=
P_t(a')-P_t(a)
+
d_{t+a}-d_{t+a'} ,
\end{aligned}
\]
which proves Eq.~\eqref{eq:incremental_return_decomposition}.
Taking conditional expectation given $\Omega_t=\omega$ and using
the definition of $V_t$ in Eq.~\eqref{eq:expected_continuation_return}
gives Eq.~\eqref{eq:population_incremental_return}.
\end{proof}

Lemma~\ref{lem:incremental-return} gives a trajectory-level interpretation of the return structure
used in Theorem~\ref{thm:monotone-allocation}. For the same additional-compute interval
$a\rightarrow a'$, the incremental return consists of the new frontier
advancement over that interval together with the change in the state's
distance from the attained frontier. Consequently, at a fixed latent
depth $t$, the increasing-differences condition takes the form
\[
\omega'>\omega
\quad\Longrightarrow\quad
\mathbb{E}\!\left[
R_t(a,a')\mid\Omega_t=\omega'
\right]
\geq
\mathbb{E}\!\left[
R_t(a,a')\mid\Omega_t=\omega
\right].
\]
Thus, the condition states that the same additional latent-compute
interval yields no smaller expected signed incremental return when more
constructive progress remains.

\begin{appcorollary}[Return ordering by remaining progress]
\label{cor:return-ordering}
Suppose that $V_t(a,\omega)$ has increasing differences in
$(a,\omega)$. For any feasible additional latent budget
$a\in\{0,\ldots,K-t\}$ and any $\omega'>\omega$,
\[
V_t(a,\omega')
\geq
V_t(a,\omega).
\]
\end{appcorollary}

\begin{proof}
By increasing differences, comparing the additional budgets $0$ and
$a$ gives
\[
V_t(a,\omega')-V_t(0,\omega')
\geq
V_t(a,\omega)-V_t(0,\omega).
\]
Since $V_t(0,\omega')=V_t(0,\omega)=0$, it follows that
\[
V_t(a,\omega')
\geq
V_t(a,\omega),
\]
which proves the result.
\end{proof}

Hence, under a common additional latent budget, larger remaining
progress corresponds to no smaller expected signed reasoning gain.

\section{Additional Experiments}
\label{app:Additional Experiments}

\subsection{Marginal Continuation Returns}
\label{app:marginal_return_diagnostic}

Appendix~\ref{app:marginal_returns_C} characterizes increasing differences through the signed marginal return over a common interval of additional latent computation. Here, we examine this return ordering beyond the one-step comparison in Table~\ref{tab:5.3}, using the same lower $\Omega_t$ and higher $\Omega_t$ groups and signed future gains. For each latent depth, we compare the marginal return of the two groups over subsequent adjacent compute intervals.

\begin{table}[htbp]
\centering
\setlength{\abovecaptionskip}{2pt}
\setlength{\belowcaptionskip}{3pt}
\setlength{\aboverulesep}{0.3ex}
\setlength{\belowrulesep}{0.3ex}
\setlength{\tabcolsep}{3pt}
\caption{\textbf{Marginal return ordering over continued latent CoT.} Each entry reports the mean signed marginal-return gap between higher- and lower-remaining-progress trajectories over the subsequent additional compute intervals, together with the fraction of intervals with a nonnegative gap.}
\label{tab:d2_return_ordering}
\small
\begin{adjustbox}{max width=\linewidth}
\begin{tabular}{@{}llcccccc@{}}
\toprule
\multirow{2}{*}{Backbone} & \multirow{2}{*}{Method} & \multicolumn{2}{c}{GSM8K} & \multicolumn{2}{c}{SVAMP} & \multicolumn{2}{c}{MultiArith} \\
\cmidrule(lr){3-4}\cmidrule(lr){5-6}\cmidrule(l){7-8}
& & \makecell{Mean return\\gap (pp)} & \makecell{Nonnegative\\intervals (\%)} & \makecell{Mean return\\gap (pp)} & \makecell{Nonnegative\\intervals (\%)} & \makecell{Mean return\\gap (pp)} & \makecell{Nonnegative\\intervals (\%)} \\
\midrule
\emph{GPT-2} & CODI & +13.1 & 100 & +4.1 & 80 & +4.7 & 60 \\
 & SIM-CoT & +14.7 & 100 & +4.2 & 60 & +15.2 & 100 \\
\cmidrule{1-8}
\emph{Llama-1B} & CODI & +15.1 & 90 & +2.9 & 80 & +5.2 & 70 \\
 & SIM-CoT & +13.6 & 100 & +2.7 & 80 & +6.2 & 60 \\
\midrule
\emph{GPT-2} & Coconut & +14.5 & 100 & +6.5 & 100 & +17.9 & 80 \\
 & SIM-CoT Coconut & +20.9 & 100 & +6.9 & 100 & +4.4 & 70 \\
\bottomrule
\end{tabular}
\end{adjustbox}
\end{table}

As shown in Table~\ref{tab:d2_return_ordering}, the mean return gap is positive in all settings, and the ordering remains nonnegative for most intervals in every setting. The pattern is most consistent on GSM8K, while SVAMP and MultiArith show greater variation across intervals. These results show that higher remaining progress is generally associated with larger signed
marginal continuation returns over subsequent latent CoT.

\subsection{Additional Results for Remaining Progress and Computation Returns}
\label{app:Additional Results Remaining Progress}
Tables~\ref{tab:5.3-svamp} and~\ref{tab:5.3-multiarith} report the corresponding analyses for SVAMP and MultiArith, both of which exhibit the same overall pattern. Higher remaining constructive progress is associated with larger future gains from continued latent CoT, while larger decreases in $\Omega_t$ correspond to larger reductions in subsequent reasoning gain. The relation between remaining progress and future reasoning gains identified in the main analysis is therefore consistent across benchmarks.

\begin{table}[htbp]
\centering
\footnotesize
\setlength{\tabcolsep}{3pt}
\setlength{\gainwd}{15pt}
\setlength{\abovecaptionskip}{2pt}
\setlength{\belowcaptionskip}{3pt}
\sisetup{retain-explicit-plus}
\def\pgap{\hspace{10pt}}
\caption{\textbf{Reasoning gains across remaining constructive progress.} We report SVAMP. $\Omega_3$ means the remaining constructive progress after 3 latent steps. Panel \textbf{(a)} reports task accuracy at Steps 3 and 4 under lower and higher $\Omega_3$. Panel \textbf{(b)} reports future reasoning gain, measured by the change in $\hat{G}_t$ from Step 3 to Steps 4 and 5, under lower and higher $\Omega_3$, with subscripts indicating the difference relative to lower $\Omega_3$. Panel \textbf{(c)} reports the change in future reasoning gain as $\Omega_t$ decreases, with negative values indicating a decrease.}
\label{tab:5.3-svamp}
\begin{adjustbox}{max width=\linewidth}
\begin{tabular}{@{}ll@{\pgap}S[table-format=2.1]S[table-format=2.1]@{}lS[table-format=2.1]S[table-format=2.1]@{}l@{\pgap}S[table-format=+1.1]S[table-format=+1.1]S[table-format=+2.1]@{}lS[table-format=+2.1]@{}l@{\pgap}S[table-format=+2.1]S[table-format=+2.1]S[table-format=+2.1]@{}}
\toprule
&&\multicolumn{6}{c@{\pgap}}{(a) Accuracy (\%)}&\multicolumn{6}{c@{\pgap}}{(b) Future gain}&\multicolumn{3}{c@{}}{(c) Future gain change}\\
\cmidrule(l{2pt}r{12pt}){3-8}\cmidrule(l{2pt}r{12pt}){9-14}\cmidrule(l{2pt}r{0pt}){15-17}
&&\multicolumn{3}{c}{Lower $\Omega_3$}&\multicolumn{3}{c@{\pgap}}{Higher $\Omega_3$}&\multicolumn{2}{c}{Lower $\Omega_3$}&\multicolumn{4}{c@{\pgap}}{Higher $\Omega_3$}&\multicolumn{3}{c@{}}{Decrease in $\Omega_t$}\\
\cmidrule(l{2pt}r{2pt}){3-5}\cmidrule(l{2pt}r{12pt}){6-8}\cmidrule(l{2pt}r{2pt}){9-10}\cmidrule(l{2pt}r{12pt}){11-14}\cmidrule(l{2pt}r{0pt}){15-17}
\multicolumn{2}{@{}l}{Backbone / Method}&{Step 3}&{Step 4}&&{Step 3}&{Step 4}&&{Step 4}&{Step 5}&{Step 4}&&{Step 5}&&{Lower}&{Middle}&{Higher}\\
\midrule
\multicolumn{17}{@{}l}{\textit{CODI family}}\\
\textit{GPT-2}&CODI&45.2&44.1&\gdn{1.1}&51.7&73.3&\gup{21.6}&-0.9&+0.5&+12.3&\gup{13.2}&+21.2&\gup{20.7}&-1.9&-9.3&-39.4\\
&SIM-CoT&37.5&36.4&\gdn{1.1}&26.5&70.6&\gup{44.1}&+0.3&-0.6&+31.7&\gup{31.4}&+40.8&\gup{41.4}&-2.1&-6.1&-38.5\\
\addlinespace[3pt]
\textit{Llama-1B}&CODI&53.3&53.3&\gz&42.6&48.9&\gup{6.3}&+1.0&+1.6&+2.3&\gup{1.3}&+15.4&\gup{13.8}&-0.6&-3.5&-22.9\\
&SIM-CoT&61.8&61.8&\gz&73.0&75.7&\gup{2.7}&-0.4&-0.7&+6.5&\gup{6.9}&+7.8&\gup{8.5}&-1.9&-7.2&-26.2\\
\midrule
\multicolumn{17}{@{}l}{\textit{Coconut family}}\\
\textit{GPT-2}&Coconut&27.0&28.8&\gup{1.8}&46.3&65.3&\gup{19.0}&+0.7&+0.1&+19.2&\gup{18.5}&+26.7&\gup{26.6}&+0.9&-0.1&-23.5\\
&SIM-CoT Coconut&42.4&42.4&\gz&41.9&56.5&\gup{14.6}&+1.3&+2.4&+14.3&\gup{13.0}&+20.2&\gup{17.8}&+1.0&-6.8&-52.9\\
\bottomrule
\end{tabular}
\end{adjustbox}
\end{table}

\begin{table}[htbp]
\centering
\footnotesize
\setlength{\tabcolsep}{3pt}
\setlength{\gainwd}{15pt}
\setlength{\abovecaptionskip}{2pt}
\setlength{\belowcaptionskip}{3pt}
\sisetup{retain-explicit-plus}
\def\pgap{\hspace{10pt}}
\caption{\textbf{Reasoning gains across remaining constructive progress.} We report MultiArith. $\Omega_3$ means the remaining constructive progress after 3 latent steps. Panel \textbf{(a)} reports task accuracy at Steps 3 and 4 under lower and higher $\Omega_3$. Panel \textbf{(b)} reports future reasoning gain, measured by the change in $\hat{G}_t$ from Step 3 to Steps 4 and 5, under lower and higher $\Omega_3$, with subscripts indicating the difference relative to lower $\Omega_3$. Panel \textbf{(c)} reports the change in future reasoning gain as $\Omega_t$ decreases, with negative values indicating a decrease.}
\label{tab:5.3-multiarith}
\begin{adjustbox}{max width=\linewidth}
\begin{tabular}{@{}ll@{\pgap}S[table-format=2.1]S[table-format=2.1]@{}lS[table-format=2.1]S[table-format=2.1]@{}l@{\pgap}S[table-format=+1.1]S[table-format=+1.1]S[table-format=+2.1]@{}lS[table-format=+2.1]@{}l@{\pgap}S[table-format=+2.1]S[table-format=+2.1]S[table-format=+2.1]@{}}
\toprule
&&\multicolumn{6}{c@{\pgap}}{(a) Accuracy (\%)}&\multicolumn{6}{c@{\pgap}}{(b) Future gain}&\multicolumn{3}{c@{}}{(c) Future gain change}\\
\cmidrule(l{2pt}r{12pt}){3-8}\cmidrule(l{2pt}r{12pt}){9-14}\cmidrule(l{2pt}r{0pt}){15-17}
&&\multicolumn{3}{c}{Lower $\Omega_3$}&\multicolumn{3}{c@{\pgap}}{Higher $\Omega_3$}&\multicolumn{2}{c}{Lower $\Omega_3$}&\multicolumn{4}{c@{\pgap}}{Higher $\Omega_3$}&\multicolumn{3}{c@{}}{Decrease in $\Omega_t$}\\
\cmidrule(l{2pt}r{2pt}){3-5}\cmidrule(l{2pt}r{12pt}){6-8}\cmidrule(l{2pt}r{2pt}){9-10}\cmidrule(l{2pt}r{12pt}){11-14}\cmidrule(l{2pt}r{0pt}){15-17}
\multicolumn{2}{@{}l}{Backbone / Method}&{Step 3}&{Step 4}&&{Step 3}&{Step 4}&&{Step 4}&{Step 5}&{Step 4}&&{Step 5}&&{Lower}&{Middle}&{Higher}\\
\midrule
\multicolumn{17}{@{}l}{\textit{CODI family}}\\
\textit{GPT-2}&CODI&87.5&87.5&\gz&61.5&76.9&\gup{15.4}&0.0&+0.4&+24.3&\gup{24.3}&+41.8&\gup{41.4}&-2.2&-28.8&-64.7\\
&SIM-CoT&87.2&87.2&\gz&2.7&94.6&\gup{91.9}&+1.0&+1.1&+83.3&\gup{82.3}&+86.7&\gup{85.6}&-0.3&-7.9&-60.2\\
\addlinespace[3pt]
\textit{Llama-1B}&CODI&85.7&85.7&\gz&75.0&75.0&\gz&0.0&-0.9&+3.1&\gup{3.1}&+32.8&\gup{33.7}&-1.3&-6.4&-27.2\\
&SIM-CoT&84.6&84.6&\gz&85.7&85.7&\gz&-0.5&-0.7&+2.2&\gup{2.7}&+0.9&\gup{1.6}&-2.5&-9.0&-42.0\\
\midrule
\multicolumn{17}{@{}l}{\textit{Coconut family}}\\
\textit{GPT-2}&Coconut&50.0&51.8&\gup{1.8}&10.6&36.2&\gup{25.6}&+9.7&+13.2&+19.4&\gup{9.7}&+77.7&\gup{64.5}&+7.6&-6.0&-56.3\\
&SIM-CoT Coconut&92.3&88.5&\gdn{3.8}&62.5&100.0&\gup{37.5}&0.0&+2.2&+35.7&\gup{35.7}&+37.5&\gup{35.3}&+20.2&-22.2&-84.4\\
\bottomrule
\end{tabular}
\end{adjustbox}
\end{table}

\begin{table}[htbp]
\centering
\footnotesize
\setlength{\tabcolsep}{3pt}
\setlength{\gainwd}{15pt}
\setlength{\abovecaptionskip}{2pt}
\setlength{\belowcaptionskip}{3pt}
\sisetup{retain-explicit-plus}
\def\pgap{\hspace{10pt}}
\caption{\textbf{Future reasoning gains under estimated remaining progress.} $f_\phi(X_3)$ denotes the estimated remaining progress after 3 latent steps. Panel \textbf{(a)} reports task accuracy at Steps 3 and 4 under lower and higher $f_\phi(X_3)$. Panel \textbf{(b)} reports future reasoning gain, measured by the change in $\hat{G}_t$ from Step 3 to Steps 4 and 5, under lower and higher $f_\phi(X_3)$, with subscripts indicating the difference relative to lower $f_\phi(X_3)$.}
\label{tab:app_estimator_bridge}
\begin{adjustbox}{max width=\linewidth}
\begin{tabular}{@{}lll@{\pgap}S[table-format=2.1]S[table-format=2.1]@{}lS[table-format=2.1]S[table-format=2.1]@{}l@{\pgap}S[table-format=+1.1]S[table-format=+1.1]S[table-format=+2.1]@{}lS[table-format=+2.1]@{}l@{}}
\toprule
&&&\multicolumn{6}{c@{\pgap}}{(a) Accuracy (\%)}&\multicolumn{6}{c@{}}{(b) Future gain}\\
\cmidrule(l{2pt}r{12pt}){4-9}\cmidrule(l{2pt}r{0pt}){10-15}
&&&\multicolumn{3}{c}{Lower $f_\phi(X_3)$}&\multicolumn{3}{c@{\pgap}}{Higher $f_\phi(X_3)$}&\multicolumn{2}{c}{Lower $f_\phi(X_3)$}&\multicolumn{4}{c@{}}{Higher $f_\phi(X_3)$}\\
\cmidrule(l{2pt}r{2pt}){4-6}\cmidrule(l{2pt}r{12pt}){7-9}\cmidrule(l{2pt}r{2pt}){10-11}\cmidrule(l{2pt}r{0pt}){12-15}
Dataset&Backbone&Method&{Step 3}&{Step 4}&&{Step 3}&{Step 4}&&{Step 4}&{Step 5}&{Step 4}&&{Step 5}&\\
\midrule
GSM8K&\textit{GPT-2}&CODI&46.4&47.6&\gup{1.2}&27.1&24.6&\gdn{2.5}&+0.2&+0.3&-0.6&\gdn{0.8}&+13.0&\gup{12.7}\\
&&SIM-CoT&43.6&44.8&\gup{1.2}&21.6&38.3&\gup{16.7}&+0.1&-0.1&+16.4&\gup{16.3}&+29.2&\gup{29.3}\\
\addlinespace[2pt]
&\textit{Llama-1B}&CODI&61.1&61.4&\gup{0.3}&36.2&37.1&\gup{0.9}&-0.3&+0.9&0.0&\gup{0.3}&+16.1&\gup{15.2}\\
&&SIM-CoT&61.1&61.6&\gup{0.5}&39.4&47.8&\gup{8.4}&+0.2&+0.1&+7.3&\gup{7.1}&+17.5&\gup{17.4}\\
\midrule
SVAMP&\textit{GPT-2}&CODI&53.0&53.6&\gup{0.6}&31.2&32.1&\gup{0.9}&-0.1&-0.4&+0.6&\gup{0.7}&+0.1&\gup{0.5}\\
&&SIM-CoT&53.3&53.0&\gdn{0.3}&23.7&27.0&\gup{3.3}&-1.0&-1.7&+2.3&\gup{3.3}&+2.6&\gup{4.3}\\
\addlinespace[2pt]
&\textit{Llama-1B}&CODI&78.4&78.4&\gz&36.6&37.5&\gup{0.9}&-0.3&-0.4&+0.2&\gup{0.5}&+0.8&\gup{1.2}\\
&&SIM-CoT&76.0&76.0&\gz&49.5&49.2&\gdn{0.3}&0.0&-0.5&+0.3&\gup{0.3}&-0.5&\gz\\
\midrule
MultiArith&\textit{GPT-2}&CODI&96.7&96.7&\gz&86.7&88.3&\gup{1.6}&-0.3&+0.1&+3.7&\gup{4.0}&+5.3&\gup{5.2}\\
&&SIM-CoT&77.8&84.4&\gup{6.6}&55.6&97.8&\gup{42.2}&+7.5&+9.0&+45.5&\gup{38.0}&+46.4&\gup{37.4}\\
\addlinespace[2pt]
&\textit{Llama-1B}&CODI&100.0&100.0&\gz&93.3&93.3&\gz&-1.0&+0.1&+0.3&\gup{1.3}&+3.0&\gup{2.9}\\
&&SIM-CoT&98.3&98.3&\gz&95.0&96.7&\gup{1.7}&+0.1&+0.2&-0.5&\gdn{0.6}&-1.2&\gdn{1.4}\\
\bottomrule
\end{tabular}
\end{adjustbox}
\end{table}

\newcommand{\dneg}[1]{\textsubscript{\textcolor{red!85!black}{$-$#1}}}

\begin{table}[t]
\centering
\caption{\textbf{Remaining-progress targets under matched latent computation.} Each comparison keeps the trajectory representation, regressor, offline trajectories, and adaptive controller fixed, and differs only in the regression target. Panel \textbf{(a)} compares regression targets under the default design, with average latent computation matched at 4.49 steps. Panels \textbf{(b)}--\textbf{(d)} compare remaining progress with the current reasoning value under alternative designs. We report final task accuracy on GSM8K averaged over the four CODI-family reasoners, with subscripts indicating the difference relative to remaining progress.}
\label{tab:target-ablation}
\small
\begin{tabular*}{\textwidth}{@{\extracolsep{\fill}}l ccccc@{}}
\toprule
 & \multicolumn{5}{c}{(a) Regression target} \\
\cmidrule(lr){2-6}
 & Remaining progress & Future gain & Current value & Shuffled & Signal free \\
\midrule
Accuracy & 47.2 & 46.9\dneg{0.3} & 45.3\dneg{1.9} & 45.1\dneg{2.1} & 45.1\dneg{2.1} \\
\end{tabular*}
\begin{tabular*}{\textwidth}{@{\extracolsep{\fill}}l ccc l ccc@{}}
\midrule
 & \multicolumn{3}{c}{(b) Representation \& (c) Regressor} & & \multicolumn{3}{c}{(d) Horizon adaptation} \\
\cmidrule(lr){2-4}\cmidrule(lr){6-8}
 & Remaining & Current & Avg. & & Remaining & Current & Avg. \\
Variant & progress & value & steps & Variant & progress & value & steps \\
\midrule
w/o latent & 47.0 & 45.2\dneg{1.8} & 4.48 & Min-max & 46.8 & 45.7\dneg{1.0} & 4.48 \\
w/o stats & 46.8 & 45.0\dneg{1.8} & 4.49 & Binary & 47.5 & 43.6\dneg{4.0} & 4.48 \\
Step $t$ only & 47.0 & 45.1\dneg{1.9} & 4.49 & Determ. & 47.2 & 45.4\dneg{1.8} & 4.49 \\
\cmidrule(lr){1-4}
Linear & 46.9 & 45.2\dneg{1.8} & 4.48 & $\rho_0=0.25$ & 45.5 & 42.2\dneg{3.3} & 3.79 \\
Shallow & 47.0 & 45.0\dneg{2.0} & 4.49 & $\rho_0=0.75$ & 48.1 & 47.3\dneg{0.8} & 5.17 \\
\bottomrule
\end{tabular*}
\end{table}

\subsection{Estimated Remaining Progress and Future Reasoning Gains}
Section~\ref{sec:5.3} shows that more remaining constructive progress corresponds to larger reasoning gains from continued latent CoT. Here, we examine whether this relation is retained when remaining progress is estimated from the current latent trajectory. Using out-of-sample estimates $f_\phi(X_3)$, we form lower progress and higher progress groups and evaluate the same reasoning gain and task accuracy outcomes.

As shown in Table~\ref{tab:app_estimator_bridge}(b), the higher estimate group obtains larger future reasoning gains at Step~4 and Step~5, with average gaps of $5.9$ and $10.4$ percentage points, respectively. Table~\ref{tab:app_estimator_bridge}(a) shows the same pattern in task accuracy, with the higher estimate group improving more from Step~3 to Step~4. In several settings, we observe that the higher estimate group starts from lower accuracy at Step~3 yet gains more after continuation, consistent with remaining progress characterizing future reasoning opportunity. Together, these results show that estimated remaining progress preserves the future gain separation identified in Section~\ref{sec:5.3}, indicating that the current latent trajectory carries information relevant to future continuation opportunity.

\subsection{Remaining-Progress Targets and Estimator Design}
\label{app:target-ablation}

Section~\ref{sec:signals} compares continuation signals under the same controller. Here, we examine whether the gain of adaptive inference comes from the remaining-progress target itself. Specifically, we keep the trajectory representation, the regressor, the offline trajectories, and the adaptive controller fixed, and substitute the regression target used to fit $f_\phi$. We consider remaining progress $\hat{\Omega}_t$, future reasoning gain $\hat{G}_K-\hat{G}_t$, the current reasoning value $-\hat{G}_t$, and a shuffled target that permutes $\hat{\Omega}_t$ across the offline trajectories, together with signal free allocation.

As shown in Table~\ref{tab:target-ablation}(a), remaining progress yields higher accuracy than the current reasoning value by 1.9 percentage points, while future reasoning gain achieves similar accuracy. If the gains came from offline supervision alone, we would expect the shuffled target to outperform signal free allocation. However, we find that it performs on par with signal free allocation, while remaining progress improves average accuracy by 2.1 percentage points under the same controller.

Table~\ref{tab:target-ablation}(b)--(d) further vary the trajectory representation, the regressor, and horizon adaptation, one component at a time. For the trajectory representation, we remove the latent state (w/o latent) or the predictive statistics (w/o stats) from $X_t$, or use only the observation at step $t$ (Step $t$ only). For the regressor, we replace the gradient-boosting regressor with a linear regressor (Linear) or a shallower gradient-boosting regressor (Shallow). For horizon adaptation, we replace $F_{\mathrm{ref}}$ with min-max scaling of $f_\phi(X_t)$ (Min-max), replace $g$ with a threshold rule that either continues to $K$ or stops at the current step (Binary), replace randomized rounding with rounding to the nearest integer (Determ.), and vary $\rho_0$. Remaining progress yields higher accuracy than the current reasoning value in all variants, with gaps from 0.8 to 4.0 percentage points.

In short, we show that the gain of adaptive inference actually comes from the remaining-progress target, which is grounded in the latent CoT dynamics of Section~\ref{sec:3}.

\end{document}